\documentclass[journal]{IEEEtran}
\usepackage{url}
\usepackage{times}
\usepackage{epsfig}
\usepackage{graphicx}
\usepackage{amsmath}
\usepackage{amssymb}
\usepackage{multirow}
\usepackage{booktabs}
\usepackage{setspace}
\usepackage{color}
\usepackage{algorithm}
\usepackage{algpseudocode}
\usepackage{algorithmicx}
\usepackage{amsmath}
\usepackage[table,xcdraw]{xcolor}
\usepackage[normalem]{ulem}
\usepackage{threeparttable}

\usepackage[square,sort,sort&compress,numbers]{natbib}
\useunder{\uline}{\ul}{}
\ifCLASSINFOpdf
\else
\fi
\begin{document}
\title{Fingerprint Presentation Attack Detection by Channel-wise Feature Denoising}
\author{
Feng Liu, Zhe Kong, Haozhe Liu, Wentian Zhang, Linlin Shen$^*$
\thanks{The correspondence author is Linlin Shen and the email: llshen@szu.edu.cn}
\thanks{Feng Liu, Zhe Kong, Haozhe Liu, Wentian Zhang, and Linlin Shen are with the College of Computer Science and Software Engineering, Shenzhen University, Shenzhen 518060, China;
SZU Branch, Shenzhen Institute of Artificial Intelligence and Robotics for Society, China;
Guangdong Key Laboratory of Intelligent Information Processing, Shenzhen University, Shenzhen 518060, China.}
}
\maketitle

\begin{abstract}
Due to the diversity of attack materials, fingerprint recognition systems (AFRSs) are vulnerable to malicious attacks.
It is thus important to propose effective fingerprint presentation attack detection (PAD) methods for the safety and reliability of AFRSs.
However, current PAD methods often exhibit poor robustness under new attack types settings. This paper thus proposes a novel channel-wise feature denoising fingerprint PAD (CFD-PAD) method by handling the redundant noise information ignored in previous studies.
The proposed method learns important features of fingerprint images by weighing the importance of each channel and identifying discriminative channels and “noise" channels.
Then, the propagation of “noise" channels is suppressed in the feature map to reduce interference.
Specifically, a PA-Adaptation loss is designed to constrain the feature distribution to make the feature distribution of live fingerprints more aggregate and that of spoof fingerprints more disperse.
Experimental results evaluated on the LivDet 2017 dataset showed that the proposed CFD-PAD can achieve 2.53\% average classification error (ACE) and a 93.83\% true detection rate when the false detection rate equals 1.0\% (TDR@FDR=1\%). Also, the proposed method markedly outperforms the best single-model-based methods in terms of ACE (2.53\% vs. 4.56\%) and TDR@FDR=1\%(93.83\% vs. 73.32\%), which demonstrates its effectiveness.
Although we have achieved a comparable result with the state-of-the-art multiple-model-based methods, there still is an increase in TDR@FDR=1\% from 91.19\% to 93.83\%. In addition, the proposed model is simpler, lighter and more efficient and has achieved a 74.76\% reduction in computation time compared with the state-of-the-art multiple-model-based method.
\emph{The source code is available at https://github.com/kongzhecn/cfd-pad.}
\end{abstract}


\section{Introduction}
In recent years, with the increasing use of mobile payment, online banking, and other applications, the number of smart devices has grown exponentially \cite{chugh2018fingerprint,liu2021one,maltoni2009handbook,marcel2014handbook}. Due to security requirements, an increasing number of smart devices tend to use automated fingerprint recognition systems for personal authentication \cite{liu2021one}. However, AFRSs are vulnerable to different types of attacks, including 2D or 3D artificial fingerprints made of gelatin and silicone \cite{liu2019high,liu2021one,chugh2017fingerprint,arora2016design,arora2017gold,engelsma2018universal}. To ensure that a user has a secure environment against presentation attacks (PA), many PAD methods have been proposed \cite{goicoechea2016evaluation,chugh2019oct}. To address this threat and promote anti-spoofing techniques, a series of fingerprint LivDet competitions have been held since 2009 \cite{ghiani2017review}. Many public datasets (e.g. LivDet 2011, LivDet 2013, LivDet 2015, and LivDet 2017) are also available to evaluate the performance of fingerprint PAD methods \cite{ghiani2017review,7358776,mura2018livdet}.

For fingerprint PADs, previous studies used traditional methods to extract handcrafted features, such as fractional Fourier and curvelet transforms \cite{lee2009fake,nikam2008fingerprint,schuckers2017fingerprint,jia2007new,antonelli2006fake,xia2018novel}. However, handcrafted features are sensitive to image noise and do not achieve good performance across "unknown" or novel spoof materials that were not available during training \cite{chugh2020fingerprint,chugh2018fingerprint}.

CNN-based methods for fingerprint PADs have also been proposed. The CNN-based method can automatically extract features of different scales, which reduces human intervention and simplifies feature extraction \cite{nogueira2016fingerprint}. Compared with handcrafted feature-based methods, CNN-based methods achieve better performance \cite{menotti2015deep,nogueira2016fingerprint,chugh2018fingerprint,7358776,liu2021fingerprint}.
The rethinking model proposed by Liu et al. \cite{liu2021fingerprint} has achieved good results. In this method, a rethinking module connects global and local modules, and the final score was calculated by averaging the global and local spoof scores. However, this method uses two models for training and calculating the average scores, based on ensemble learning. This multi-model-based method requires the integration of multiple-models and makes training complex and time-consuming, which is not suitable for efficient deployment. This method also investigates the heatmap in the local module and found that there exist discriminative regions for PAD, but the authors did not explore this phenomenon.

\begin{figure}[!]
    \centering
    \includegraphics[scale=0.40]{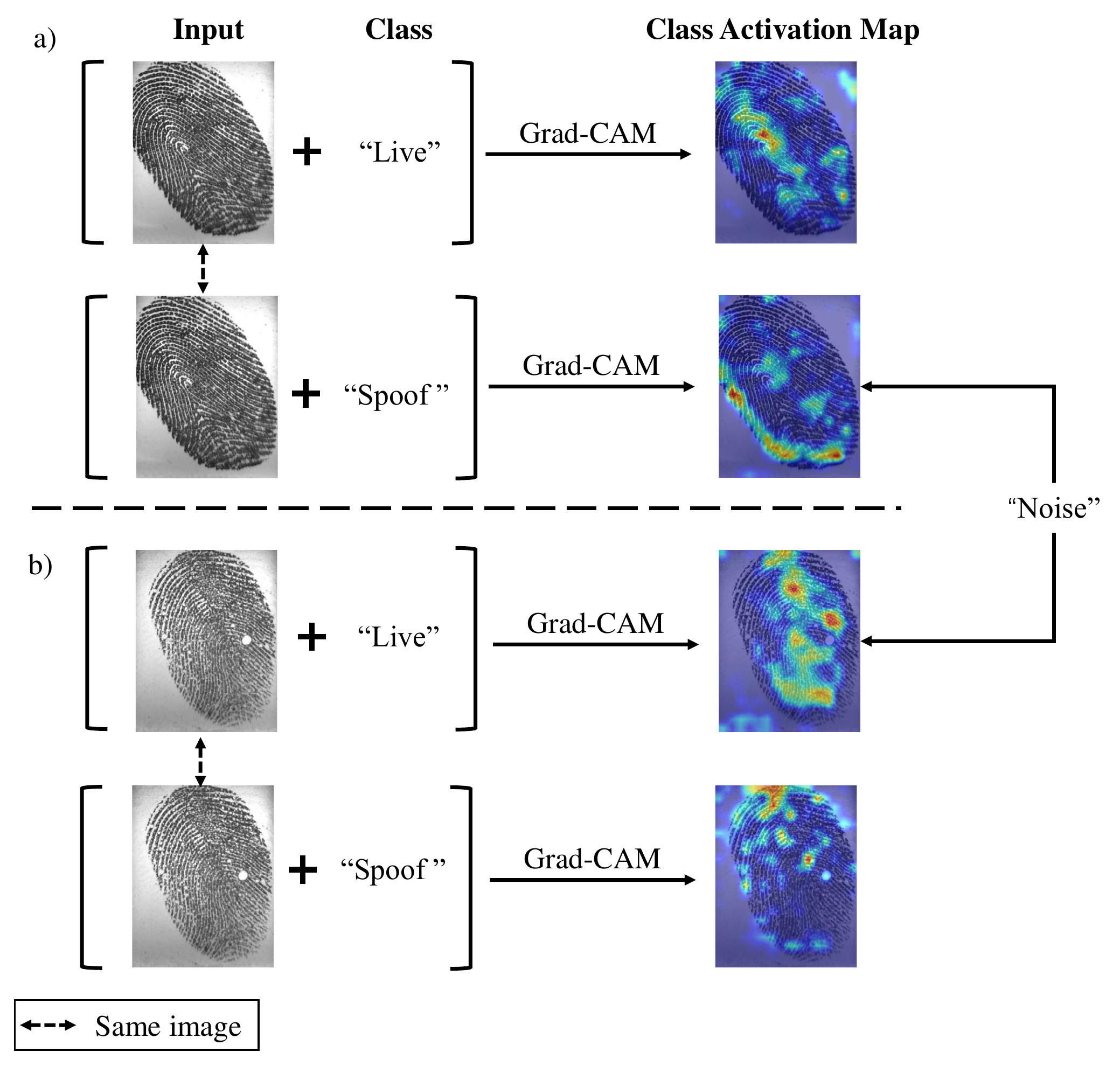}
   \caption{ Class activation maps of the samples activated by the given labels "live" and "spoof". a) A live input. b) A spoof input.}
   \label{fig:liveAndSpoof}
\end{figure}

Based on these considerations, we made a further analysis of Grad-CAM and found an interesting phenomenon. Grad-CAM \cite{selvaraju2017grad} is a visual explanation technique used to produce a coarse localization map that highlights important regions related to a given label. 

The visualised localization map $L_{Grad-CAM}^c$ is obtained by the following equation:

\begin{equation}
    L_{Grad-CAM}^c = ReLU(\sum_{k} \alpha_k^c A^k)
\end{equation}

where $c$ denotes the "class" in Fig 1, set to "live" or "spoof". $A$ is the feature map output by the convolutional layer and $A^k$ denotes the k-th channel in the feature map $A$. The neuron importance weights $\alpha_k^c $ can be calculated by:

\begin{equation}
    \alpha_k^c = \frac{1}{Z}\sum_{i}\sum_{j}\frac{\partial  y^c}{\partial  A_{ij}^k}
\end{equation}

where $y^c$ denotes the gradient of the score for class c, $A_{ij}^k$ denotes the $i$-th row $j$-th column of the k-th channel in $A$. In fact, this weight $\alpha^c_k$ represents a partial linearizatioin of $A$, and captures the 'importance' of feature map $k$ for a target class $c$. Hence, if $A$ is the feature map of the "live" fingerprint image and $c$ is "spoof" class, or $A$ is the feature map of the "spoof" fingerprint image and $c$ is "live" class, the "noise" localization map $L_{Grad-CAM}^c$ is derived.

As shown in Fig. \ref{fig:liveAndSpoof}a, given a live sample with the "live" label, Grad-CAM highlights those regions critical to identify a "live" sample; we denote those features as "live" features. When labeling the same live sample with a "spoof" label, the model also activates other regions that help to identify a "spoof" sample; we denote those features as "spoof" features. The same phenomenon appears in the spoof sample (Fig. \ref{fig:liveAndSpoof}b). Hence, as shown in Fig. \ref{fig:liveAndSpoof}, when the "Input" and "Class" have different labels, the "noise" localization map $L_{Grad-CAM}^c$ can be calculated.. Therefore, we conclude that an input image, whether live or spoof, contains both "live" and "spoof" features. But for these two types of features, they may have different effects to different fingerprint images. For a \textbf{live} sample, those regions activated by the label \textbf{"spoof"} may mislead the model and shall be identified as "noise". For a \textbf{spoof} sample, those regions activated by the label \textbf{"live"} are also able to fool the model and increase the probability of the "live" class. As each channel (i.e. feature vector) in CNN is extracted by a learned filter, we propose a novel fingerprint PAD method based on channel-wise feature denoising to remove those noises.

The remainder of this paper is organized as follows. In Section II, we provide a brief overview of other methods related to fingerprint PAD. Section III introduces the proposed method in details. Then, Section IV shows the empirical evaluation of the proposed method in LivDet and comparisons with other PAD methods. Finally, we conclude the paper in Section V.

\section{Related Work}

Because the framework of the proposed PAD method is different from current PAD methods, we review the literature of current PAD methods and the feature denoising methods used in computer vision in this section.

Typically, PAD methods can often be divided into two categories: hardware and software-based approaches \cite{chugh2019oct,marasco2014survey}. In hardware approaches, special sensors such as imaging technologies based on multispectral, shortwave infrared and optical coherent tomography (OCT), have been developed to extract live features such as odor, blood flow and heartbeat to discriminate live samples from spoof samples \cite{robison2019system,tolosana2018towards,tolosana2019biometric,moolla2019optical,baldisserra2006fake,nixon2004novel,reddy2008new,liu2019high,agassy2019liveness,liu2021one,liu2020flexible,engelsma2018raspireader}. For example, a spectroscopy-based device proposed by Nixon et al. \cite{nixon2004novel} can obtain the spectral features of the fingertip, which has been shown to be able to effectively distinguish between spoof and live fingertips. Reddy et al. \cite{reddy2008new} used a pulse-oximetry-based device to capture the percentage of oxygen in the blood and the pulse rate for live detection. OCT devices were applied by Liu et al. \cite{liu2019high} to collect depth-double-peak features and subsingle-peak features of fingerprints, which achieved 100$\%$ accuracy over four types of artificial fingerprints. Ultrasonic fingerprint sensors were used by Agassy et al. \cite{agassy2019liveness} to determine whether a fingertip is spoof or live. Liu et al. \cite{liu2021one} proposed a one-class framework for PAD using OCT. Because only live fingerprints are required for training, this method has better generalization capability and alleviates data dependence issues.

Software-based methods do not require any additional hardware devices and can also be divided into two categories: traditional PAD methods and CNN-based PAD methods. Traditional methods distinguish spoof from live fingerprints by extracting handcrafted features such as anatomical features (e.g. the locations and distribution of core), physiological features (e.g. perspiration and ridge distortion), and texture-based features from the fingerprint images \cite{schuckers2017fingerprint,jia2007new,antonelli2006fake,xia2018novel}. For example, skin distortion information, fractional Fourier transforms and curvelet transform-based methods are all effective methods used to extract handcrafted features to discriminate live and spoof fingerprints \cite{antonelli2006fake,lee2009fake,nikam2008fingerprint}.

With the development of computer vision, CNNs have begun to be widely used in fingerprint PADs. CNN-based methods are learning-based, and the convolutional layer with a pooling layer in CNN can automatically extract features of different scales. Compared with traditional methods, CNN-based methods reduce human intervention and make feature extraction simple because there is no need to manually design feature generators for specific tasks \cite{nogueira2016fingerprint}. CNN-based methods outperform solutions of handcrafted features by a wide margin \cite{menotti2015deep,nogueira2016fingerprint,chugh2018fingerprint,7358776}.
For example, Nogueira et al. \cite{nogueira2014evaluating} implement and evaluate two different feature extraction techniques: convolutional networks with random weights and local binary patterns, for PAD. Then, a support vector machine (SVM) is used for classification. In \cite{nogueira2016fingerprint}, Nogueira et al. first pretrain deep CNNs for object recognition and then fine-tuned the CNN for fingerprint PAD. Chugh et al. \cite{chugh2018fingerprint} proposed a deep convolutional neural network-based method. In their method, they used fingerprint prior knowledge by extracting local patches centered and aligned using minutiae to train the MobileNet-v1 CNN model.

The methods in \cite{nogueira2016fingerprint,chugh2018fingerprint,nogueira2014evaluating} improved the performance of fingerprint PAD. However, their generalization capability is poor. Improving generalization performance across "unknown" or novel attacks is essential for fingerprint PAD. Marasco et al. \cite{marasco2011robustness} pioneered the study of unknown attacks in fingerprint PAD. 
In \cite{rattani2014automatic}, the authors designed an PAD scheme automatic adaptation to detect novel unknown attacks. One-class based approach is also used to address novel unknown attacks problem. In \cite{sequeira2015fingerprint}, the authors compared the performance of supervised and semisupervised approaches that rely solely on the bona fide samples. Results indicated the vulnerability of the biometric systems. To improve generalization performance against novel attacks, Menotti et al. \cite{pereira2020robust} used adversarial training and representation learning to improve the generalization capacity of PAD model to the unseen attack. In addition, a GAN-based method is also used.
Chugh et al. \cite{chugh2020fingerprint} proposed a novel style-transfer-based method to improve both cross-material and cross-sensor robustness of fingerprint PADs. They transferred the style characteristics between fingerprint images of known materials to synthesize fingerprint images corresponding to unknown materials. In this method, they used centered and aligned minutiae-based prior knowledge to propose a style-transfer-based wrapper, which is used to transfer the style characteristics of known fingerprint materials. Then, the CNN was trained to learn generative-noise invariant features to improve generalization capability. This method is based on data augmentation, which increases the ACE of Slim-ResCNN and FSB from 94.01\% and 95.44\% to 95.37\% and 95.88\%, respectively \cite{zhang2019slim,chugh2018fingerprint,chugh2020fingerprint}.

Then, Liu et al. \cite{liu2021fingerprint} proposed a global-local model-based PAD method. In this method, no images of the target material or target sensor appear in the training set, and they proposed a rethinking module to connect global and local modules. The final spoof score is calculated by averaging the global and local spoof scores. This method achieved an average classification error of 2.28\% in cross-sensor settings. However, this method is based on ensemble learning, which requires complex and time-consuming training of multiple models. Though achieved good performance using global-local features, the approach is time-consuming and thus not suitable for real-word applications.
The authors also showed some discriminative regions for PAD, but they did not further explore this phenomenon. In this paper, we explored the "noise" regions in a fingerprint image, and propose a feature denoising method to improve the generalization capability of fingerprint PAD.

\begin{figure*}[!]
    \centering
        \includegraphics[scale=0.56]{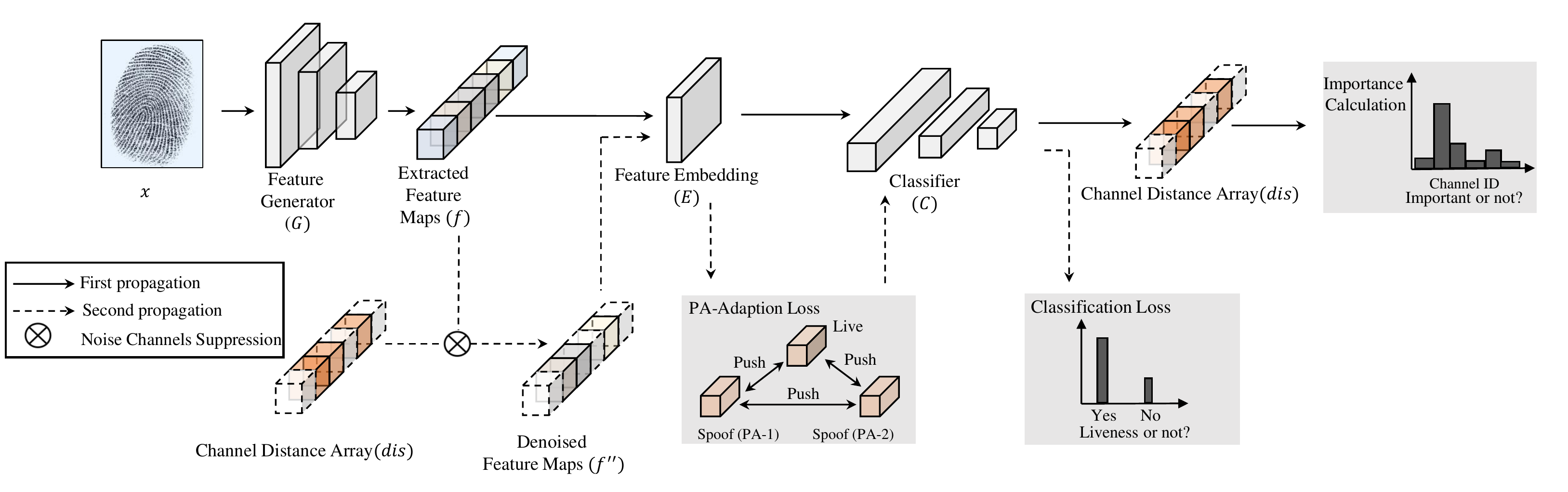}
   \caption{An overview of the proposed fingerprint PAD method(CFD-PAD) based on channel-wise feature denoising.}
    \label{fig:pipeline}
\end{figure*}

Feature denoising is a common method in computer vision. Traditional denoising methods often denoise through image analysis and modeling \cite{brooks2019unprocessing}. Many classic methods such as anisotropic diffusion and total variation denoising, use custom algorithms to recover a clean signal from noise input under the assumption that both signal and noise exhibit particular statistical regularities \cite{perona1990scale,rudin1992nonlinear,brooks2019unprocessing,zhang2017beyond}. The traditional denoising method is simple and effective, but the parametric models limit their capacity and expressiveness \cite{brooks2019unprocessing}. However, the essence of the current feature denoising in the deep learning-based method is to suppress the "noise" in the feature map to make the responses focus on visually meaningful regions \cite{xie2019feature}. For example, in the field of adversarial attacks, Xie et al. \cite{xie2019feature} found that adversarial perturbations on images lead to "noise" in the features extracted by convolutional networks. Thus, they developed new network architectures that increase adversarial robustness by performing feature denoising. Specifically, the networks contain blocks that denoise the features using nonlocal means or other filters. The feature denoising networks substantially improve the adversarial robustness in both white and black-box attack settings. 
Similar to feature denoising, attention is another useful module in computer vision \cite{vaswani2017attention,hu2018squeeze,fu2019dual,chaudhari2019attentive,xu2015show}. Attention models tend to focus selectively on some important parts of the image while ignoring other irrelevant parts \cite{chaudhari2019attentive,xu2015show}. Fu et al. \cite{fu2019dual} proposed a novel dual attention network (DANet) with a self-attention mechanism to enhance the discriminant ability of feature representations for scene segmentation. These studies show that the denoising strategy is critical for computer vision and has already been successfully applied in different tasks. However, using denoising strategies in classification tasks, particularly in PAD, has rarely been investigated. 

In this paper, for the first time, we use a denoising strategy to improve the generalization capability of fingerprint PAD. Fig. \ref{fig:pipeline} shows the flowchart of the proposed method. The training stage contains three parts: a) extracting "noise" channels from feature maps; b) suppressing those selected "noise" channels to suppress interference; and c) designing PA-Adaptation loss to align the feature domain of the fingerprint. Given an input image, the model first evaluates the importance of each channel in the feature map to find those "noise" channels, which correspond to "noise" regions with abundant "noise" features. Then, we minimize the interference of "noise" regions by suppressing the "noise" channels. Finally, to alter the feature distribution of different attack types, we design a PA-Adaptation loss to align the feature domain of the fingerprint by pulling all the live fingerprints close and pushing the spoof samples apart. In this paper, we propose a novel fingerprint PAD method based on channel-wise feature denoising. Unlike other methods, we are the first group to propose the "noise" regions in fingerprint PAD. We successfully found those channels that correspond to "noise" regions and remove the interference of "noise". Experimental results show that the proposed method is efficient and requires less computation than existing methods. The experimental results show that our approach significantly improve the performance of PAD, compared with the state-of-the-art method, in both cross-sensor and cross-material settings.

The primary contributions of this paper are summarized as follows.

\begin{itemize}
\item[$\bullet$] A channel-wise feature denoising model is first proposed for fingerprint PAD in this paper. By filtering the redundant information or “noise” regions on the feature map, the fingerprint PAD performance can be substantially improved.
\item[$\bullet$] An effective method to evaluate each channel and suppress the propagation of “noise” channels is proposed. By weighting the importance of each channel, “noise” channels can be suppressed.
\item[$\bullet$] A new loss function that alters the feature distribution of spoof types is designed. A PA-Adaptation loss is designed to align the feature domain of the fingerprint by pulling all the live fingerprints close and pushing different spoofs apart.
\item[$\bullet$] Experimental results performed on LivDet 2017 show that the proposed CFD-PAD can achieve an ACE of 2.53\% and a TDR of 93.83\% when the FDR equals 1.0\% in the case of cross-material setting. In the case of the cross-sensor setting, the proposed method achieves a 19.80\% ACE and a 32.32\% TDR when the FDR equals 1.0\%. The proposed method is based on a single lightweight model, which requires less computation time and is suitable for real applications.
\end{itemize}

\section{Proposed Approach}
As the overview shown in Fig. \ref{fig:pipeline}, the proposed method, called CFD-PAD, applies a channel-wise denoising model for PAD. The proposed method includes three stages in the training process: evaluating the importance of each channel, suppressing the propagation of "noise" channels, and aligning channel-wise domains. In the first stage, a channel distance array $dis$ is generated to measure the importance of each channel in the feature map $f$. In the second stage, we use the obtained channel distance array $dis$ to suppress the propagation of those "noise" channels. After suppressing the "noise" channels in $f$, we obtain $f^{''}$. In the last stage, PA-Adaptation loss is used to pull all the live fingerprints close while pushing the spoof ones of different attack types apart. In this section, we first provide the system framework of the proposed method and then introduce these three stages in detail.

\begin{algorithm}[!]
    \caption{Channel-wise Feature Denoising PAD Method (CFD-PAD)}
    \label{alg::cfd_pad}
    \begin{algorithmic}
    \Require
        \\ Training Set of Images:$X_{train}$=\{$x_{1}$,$\cdots$,$x_{i}$,$\cdots$,$x_{n}$\};
        \\ Groundtruth of $X_{train}$:$Y_{train}$=\{$y_1$,$\cdots$,$y_i$,$\cdots$,$y_n$\};
        \\ Attack type of $X_{train}$:$A_{train}$=\{$a_1$,$\cdots$,$a_i$,$\cdots$,$a_n$\};
        \\ Feature Generator: G(·) with learning parameters $\theta_g$;
        \\ Feature Embedding: E(·) with learning parameters $\theta_e$;
        \\ Classifier: C(·) with learning parameters $\theta_c$;
        \\ Loss Function: Cross-entropy Loss function $\mathcal{L}_{c}$(·,·) and PA-Adaptation Loss function $\mathcal{L}_{padp}$(·,·);
        \\ Training Epoch Number: $e$;
        \\ Learning Rate: $\alpha$;
        \\ Channel's number of $g_t$: $c$
        \\ channel distance array: $dis$
    \Ensure 
    \end{algorithmic}
    \begin{algorithmic}[1]
        \State Initialize $\theta_g$, $\theta_e$, $\theta_c$ from the model pretrained on ImageNet. 
        \State Set all the values in array $dis$ to 0
        \For {$j=1$ to $e$}
            \For{ $x \in X_{train}$ }
                \State $f \gets$ $G(x)$;
                \State $e \gets$ $E(f)$; 
                \State $o \gets$ $C(e)$;
                \State $r \gets$ $softmax(o)$;
                
                \State // Evaluate the importance of each channel
                \For {$i=1$ to $c$}
                    \State $f_{i}^{'} \gets$ Suppress the $i$-th channel of $f$;
                    \State $e_{i}^{'} \gets$ $E(f_{i}^{'})$;
                    \State $o_{i}^{'} \gets$ $C(e_{i}^{'})$;
                    \State $r_{i}^{'} \gets$ $softmax(o_{i}^{'})$;
                    \State $dis[i] \gets$ $dis[i]$ + $abs(a$-$a_i)$;
                \EndFor
                \State // Suppress the propagation of "noise" channels
                \For {$i=1$ to $c$}
                    \If{$dis[i]$ is in $maxk(dis)$}
                        \State  $f^{''}[i] \gets$ $f^{'}[i]$
                    \Else
                        \State $f^{''}[i] \gets$  0 
                    \EndIf
                \EndFor
                \State $e^{''} \gets$ $E(f^{''})$; 
                \State $o^{''} \gets$ $C(e^{''})$;
                
                \State $\mathcal{L}_{cfd} \gets \lambda_1\mathcal{L}_{c}(o,y) + \lambda_2\mathcal{L}_{padp}(e^{''},a) + \lambda_3\mathcal{L}_{c}(o^{''},y)$
                \State Update $\theta_g$, $\theta_e$, $\theta_c$
                
            \EndFor
        \EndFor
    \end{algorithmic}
\end{algorithm}

\begin{figure*}[!]
    \centering
        \includegraphics[scale=0.56]{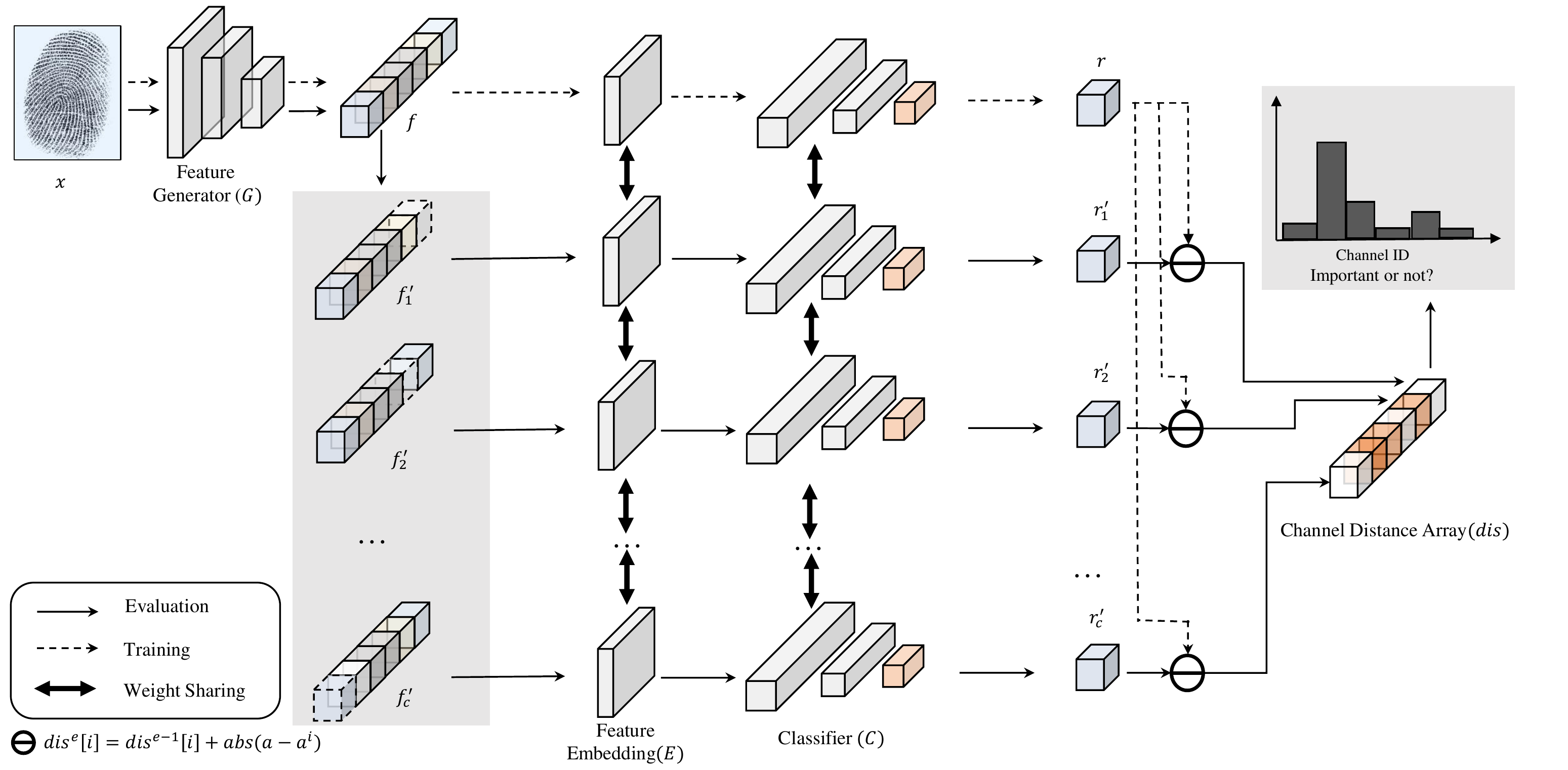}
   \caption{Illustration of weighting the importance of each channel in the feature map $f$. The feature maps $f^{'}$ of the gray background is derived from suppressing each channel in the feature map $f$. And the importance of each channel can be obtained by calculating the distance between $r$ and each element in $r_{i}^{'}$ ($i$ is range from $1$ to $c$). The output $r^{'}_{i}$ is derived after $f^{'}_{i}$ passes through the $E(\cdot)$ and $C(\cdot)$.}
    \label{fig:weightChannel}
\end{figure*}

\subsection{System Framework}
To better describe the proposed method, we first set up the basic notations used in this paper: ($x$, $y$) is a data sample, $x$ is an input image and the corresponding label is $y$. $G(\cdot)$ is the feature generator module; $E(\cdot)$ is the feature embedding module, $C(\cdot)$ is the classifier; and $\theta_g$, $\theta_e$, and $\theta_c$ are their respective parameters. We use $f$ to denote the feature map output by $G(\cdot)$, and $e$ is the feature map output by $E(\cdot)$. Therefore, we have the following equations:
\begin{equation}
    g = G(x)
    \label{eq.g}
\end{equation}
\begin{equation}
    e = E(g)
    \label{eq.e}
\end{equation}
\begin{equation}
    o = C(e)
    \label{eq.o}
\end{equation}
where $o$ denotes the prediction result of the network and we can get $r$ when using the softmax function to normalize $o$, as formulated in Eq.(\ref{eq.r}).
\begin{equation}
    r = softmax(o)
    \label{eq.r}
\end{equation}

The system primarily consists of three steps during training, as shown in Fig.\ref{fig:pipeline}: importance evaluation of each channel, propagation suppression of "noise" channels (unimportant channels) in the feature map, and channel-wise domain alignment. As described in the Algorithm. \ref{alg::cfd_pad}, attack type denotes the element and characteristic of a presentation attack, including presentation attack instrument (PAI) species, concealer or impostor attack; degree of supervision, and method of interaction with the capture device \cite{standard2017information}. We input an image $x$ and then obtain a feature map $f$ from $G(\cdot)$ by Eq. (\ref{eq.g}) in the first step. Then, a channel distance array $dis$ is generated according to the importance of each channel in the feature map $f$. In the second step, according to $dis$, we divide the channels into important channels and "noise" channels. Then we can obtain $f^{''}$ after suppressing those "noise" channels in $f$ in the channel suppression module by:
\begin{equation}
    f^{''}[i] = 
    \begin{cases}
        f[i], & dis[i] \in maxk(dis) \\
        0, & dis[i] \notin maxk(dis) \\
    \end{cases}
    \label{eq.dn}
\end{equation}
where $i$ is the i-th channel in the feature map and $maxk(\cdot)$ is the top k values in an array.
In the channel-wise domain alignment step, $e^{''}$, derived by Eq. (\ref{eq.e}), is put to the PA-Adpatation loss for gradient updating.

\subsection{Importance Calculation For Each Channel}
Calculating the importance of each channel can guide the network to distinguish between important channels and "noise" channels in $f$. A detailed description of this process is shown in Figure \ref{fig:weightChannel}.

Given an input image $x$, we first pass it into the feature generator $G(\cdot)$ to gain a feature map $f$. The derived feature map $f$ has $c$ channels, and then, we successively zero channel $i$ (range from $1$ to $c$) in the feature map $f$, which remains unchanged for the other channels. Therefore, we obtain $c$ new feature maps $\{ f^{'}_{1}$, $f^{'}_{2}$, ... ,$f^{'}_{c}$ \}. Next, we feed the original feature map $f$ and all $c$ new feature maps $\{ f^{'}_{1}$, $f^{'}_{2}$, ... ,$f^{'}_{c}$ \} into the feature embedding layer $E(\cdot)$, the classification layer $C(\cdot)$ and the softmax layer and thus obtain the corresponding representation $r$, $\{r^{'}_{1}$, $r^{'}_{2}$, ..., $r^{'}_{c}\}$. Finally, we can obtain the channel distance array $dis$ by calculating the distance between representation $r$ and $r^{'}_i$ ($i$ ranges from $1$ to $c$), where $r$ is a two-tuple matrix and $r$ = \{$a$, $b$\}. Because PAD is a binary classification task, $a$ and $b$ in $r$ denote the probability of live and spoof fingerprints, respectively, where $a$ and $b$ range from 0 to 1, and the sum of $a$ and $b$ is 1. Therefore, the distance array $dis$ can be calculated by:
\begin{equation}
    dis^{b}[i] = dis^{b-1}[i] + abs(a - a_i)
\end{equation}
where $ dis^{b}[i] $ is the importance of the i-th channel in $f$ and $b$ is the batch number.
Because the calculation of $dis$ is cumulative, $dis^{b-1}[i]$ denotes the calculation result of the top $(b-1)$ batch. When training begins, $b=1$, and $dis^{b-1}[i]$ is 0.
$abs(a - a_i)$ describes the change in the network's output caused by suppressing the i-th channel of $f$. If the channel $i$ is indeed important, the corresponding representation $r^{'}_{i}$ derived from $E(\cdot)$ and $C(\cdot)$ will change considerably compared with $r$. The distance $abs(a - a_i)$ between representations $r$ and $r^{'}_{i}$ is dynamic. The larger the value of $abs(a - a_i)$, the suppression of the i-th channel has a more significant influence on the result, which also indicates that this channel belongs to important channels. Conversely, a small value of $abs(a - a_i)$ means that the network output is not sensitive to the suppression of the i-th channel, which shows that the i-th channel is a "noise" channel. Because evaluating the importance of each channel is a dynamic accumulation process, the importance of each channel in the e-th batch is jointly determined by the current batch and the previous $(b-1)$ batch.

We calculate the importance of each channel in $f$ to obtain $c$ values $ dis = \{ dis[1], dis[2], ……, dis[k], ……, dis[c] \} $ and sort $dis$. More formally, the $maxk(\cdot)$ function means the top $k$ values in an array and we select the top $k$ channels as important channels. Correspondingly, the remaining $(c-k)$ channels are thus "noise" channels. In addition, $dis$ is a dynamically accumulated channel distance array, and we accumulate the distance between $r$ and $r^{'}_{i}$ ( $i$ ranges from $0$ to $c$) in every batch.

\subsection{Channels Suppression}
Because all the channels in $f$ are divided into important channels and "noise" channels according to $dis$, channels suppression mitigates the propagation of "noise" channels and retains important channels. After suppressing those "noise" channels in $f$, we can obtain $f^{''}$.

The channel distance array $dis$ describes the importance of each channel in the feature map $f$. $dis$ is an array containing $c$ elements, and each element in $dis$ represents the importance of the corresponding channel in $f$. $maxk()$ is used to select the top-k value in $dis$ and the index of the k values is the number ID of important channels in $f$. The remaining channels are "noise" channels. Abundant discriminative features exist in important channels and "noise" features exist in "noise" channels. To eliminate the reflection of "noise" features, we set 0 to those "noise" channels in $f$, while remaining unchanged for other important channels, as list in Eq (\ref{eq.dn}). Because the corresponding gradient will disappear after setting 0 to those "noise" channels, the goal of denoising is achieved. We obtain $f^{''}$ after denoising the feature map $f$ and the denoised feature map $f^{''}$ will replace $f$ to propagate in the network. When $f^{''}$ passes through $E(\cdot)$ and $C(\cdot)$, we can obtain $e^{''}$ and $o^{''}$ by Eq.(\ref{eq.e}) and Eq.(\ref{eq.o}).


\begin{table*}[]
\centering
\caption{SPECIFICATION FOR BACKBONE: MOBILENET V2 ARCHITECTURE \cite{2018MobileNetV2}}
\label{fig.mobilenetv2}
\setlength\tabcolsep{30pt}
\large
\resizebox{.98\textwidth}{!}{%
\begin{tabular}{c|c|c|c|c|c}
\hline
Input        & Operator    & Expansion factor & Output-Channel & Repeated times & Stride \\ \hline
224 × 224 × 3          & conv2d      & -                & 32             & 1              & 2      \\
112 × 112 × 32         & bottleneck  & 1                & 16             & 1              & 1      \\
112 × 112 × 16         & bottleneck  & 6                & 24             & 2              & 2      \\
56 × 56 × 24         & bottleneck  & 6                & 32             & 3              & 2      \\
28 × 28 × 32         & bottleneck  & 6                & 64             & 4              & 2      \\
14 × 14 × 64         & bottleneck  & 6                & 96             & 3              & 1      \\
14 × 14 × 96$ ^*$        & bottleneck  & 6                & 160            & 3              & 2      \\
7 × 7 × 160        & bottleneck  & 6                & 320            & 1              & 1      \\
7 × 7 × 320        & conv2d 1×1  & -                & 1280           & 1              & 1      \\
7 × 7 × 1280       & avgpool & -                & -              & 1              & -      \\
1 × 1 × 1280     & conv2d 1×1  & -                & 2              & -              &       \\ 
\hline
\end{tabular}%
}
\begin{tablenotes}
\footnotesize
\item[*] \setlength{\parindent}{10em} * The derive feature map $z_t$.
\end{tablenotes}
\end{table*}

\begin{figure}[!]
    \centering
        \includegraphics[scale=0.50]{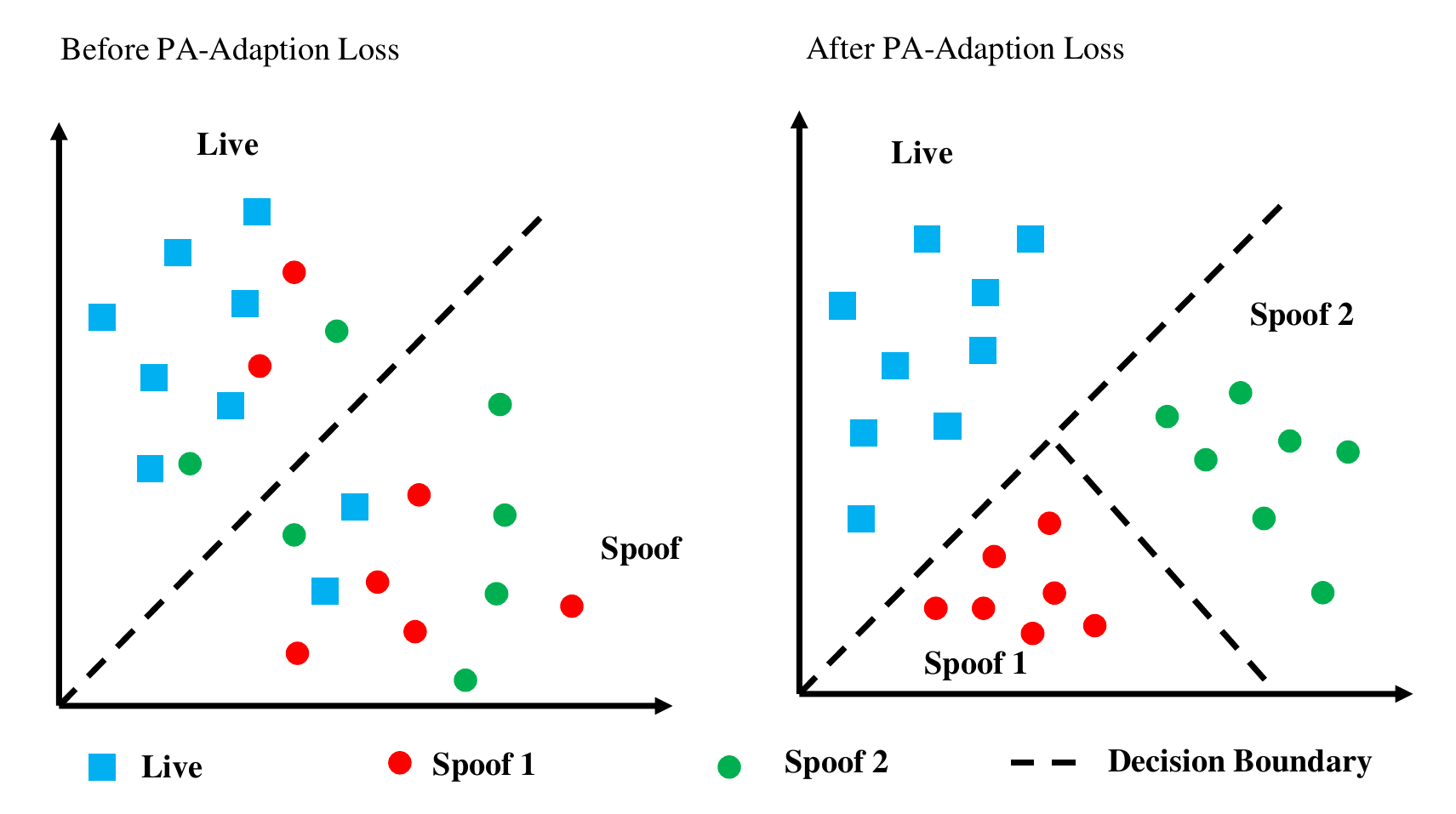}
  \caption{Description of the PA-Adaptation loss. Left: The distribution of fingerprint images before using PA-Adaptation. Right: The distribution of fingerprint images after using PA-Adaptation.}
    \label{fig:distribution}
\end{figure}

\subsection{Design of PA-Adaptation Loss}
Due to the diversity of attack types, the distribution discrepancies among spoof fingerprints are much larger than those among live fingerprints. Therefore, seeking a generalized decision boundary for fingerprints is difficult and may also affect the classification accuracy for the target domain. Thus, we proposed to align the feature domain of fingerprints by pulling all the live fingerprints close and pushing different spoof attacks apart.

As shown in Fig.\ref{fig:distribution}, the points in different shapes represent the feature distribution of different fingerprints. The square points denote the live fingerprints, and the circular points in different colors denote different types of spoof fingerprints. Because the features of live and spoof fingerprints of different attack types are randomly distributed, it is difficult to find a precise decision boundary for live and spoof fingerprints. We propose PA-Adaptation loss to alter the feature distribution by pulling all the live fingerprints close while pushing the spoof ones of different attack types apart. The left figure in Fig. \ref{fig:distribution} shows the distribution of fingerprints without using PA-Adaptation loss, and the right figure shows the distribution after using PA-Adaptation loss.
In PA-Adaptation loss, spoof fingerprints of different attack types are treated as different classes. We assume that the training set has a total of $n$ types of spoof fingerprints; thus, PA-Adaptation will push the $n+1$ class (i.e., spoof fingerprints with $n$ different attack types and 1 live fingerprint) apart but pull close the fingerprint images of the same class. Therefore, for a fingerprint image $x$ (anchor), other images belonging to the same class as $x$ are positive samples, and those belonging to other classes (other n classes) are negative samples. The intraclass distance will decrease, and the interclass distance will increase. After PA-Adaptation loss, the following optimization goals are achieved. First, the live fingerprints is clustered, and the distance between live and spoof fingerprints increases. Second, while the fingerprint of the same attack type is pulled close, fingerprints of different attack type are pushed apart. The constraint of PA-Adaptation loss is given by:
\begin{equation}
    \mathcal{L}_{padp}=max ( || E(f^a) - E(f^p) ||_2^2 - || E(f^a) - E(f^n) ||_2^2 + \alpha, 0 )
    \label{eq:triplet}
\end{equation}
where the anchor $f^a$ and the positive example $f^p$ are all feature maps with the same label, while $f^n$ is the feature map of a negative sample, which has a different label with $f^a$. $\alpha$ is a margin enforced between positive and negative pairs. $E(\cdot)$ is the feature embedding layer, and $f$ is a feature map generator by the feature generator $G(\cdot)$. Thus, Equation (\ref{eq:triplet}) can also be written as:
\begin{equation}
    \mathcal{L}_{padp}=max ( || e^a - e^p ||_2^2 - || e^a - e^n ||_2^2 + \alpha, 0 )
    \label{eq:triplet2}
\end{equation}
where $||e^a-e^p||_2^2$ is the distance between the anchor and positive sample, and $||e^a-e^n||_2^2$ is the distance between the anchor and negative sample. We minimize this loss to push $||e^a-e^p||_2^2$ to 0 and $||e^a-e^n||_2^2$ to be greater than $||e^a-e^p||_2^2+ \alpha $. As soon as $||e^a-e^p||_2^2+ \alpha < ||e^a-e^n||_2^2$, the loss becomes zero. After PA-Adaptation loss, the extracted features of fingerprints in different attack types are more dispersed than before and that of live fingerprints is aggregated, which achieves good performance for unknown attack types.

Also, to pull all the live fingerprints closer and push the spoof fingerprints of different attack types apart, we use $\mathcal{L}_{padp}(e^{''}, a)$ to calculate their PA-Adaptation loss for the gradient update.
$a$ is a more fine-grained label compared with $y$. The fine-grained label $a$ not only contains the live or false label of the input sample $x$, but also contains the attack type labels. $e^{''}$ is a denoised feature map, calculated by Eq. (\ref{eq.g}), Eq. (\ref{eq.e}) and Eq. (\ref{eq.dn}). Using the denoised feature map for PA-Adaptation loss can force the network to be less interfered by "noise" channels, which encourages the network to learn a better feature distribution for unknown attack types. 

Integrating all of these components together, the objective of the proposed channel-wise feature denoising method for fingerprint PAD is:
\begin{equation}
    \mathcal{L}_{cfd} = \lambda_1 \mathcal{L}_{c}(o, y) + \lambda_2 \mathcal{L}_{padp}(e^{''}, a) + \lambda_3 \mathcal{L}_{c}(o^{''}, y)
\end{equation}
where $\lambda1$, $\lambda2$ and $\lambda3$ are the weighting parameters. The values of $\lambda1$, $\lambda2$ and $\lambda3$ are chosen by experiment. We refer to the value ranges of multiple loss functions in the paper \cite{jia2020single} and do a lot of experiments to fine tune them. They were manually selected through experiments. And the optimal values of these three parameters for each case are different. $\mathcal{L}_{c}$ is the cross-entropy loss.

\section{Experiments And Results}
To evaluate the performance of the proposed method, several experiments were performed on the LivDet 2017 dataset \cite{mura2018livdet} in this section. This dataset is one of the most commonly used public datasets for fingerprint PAD. The results of an ablation study are shown in Table \ref{tab.ablation}. Comparative results of the cross-material setting with other methods are shown in Table \ref{tab.crossmaterial}, and the comparative results of the cross-sensor setting are shown in Table \ref{tab.crosssensor}. More details in each case are shown in Table \ref{tab.result}.

\subsection{Dataset and Implementation Details}
Table \ref{tab.livdet2017} shows a summary of the LivDet 2017 dataset, which is one of the most recent publicly available LivDet datasets and contains over 17,500 fingerprint images captured from three different scanners, Green Bit, Orcanthus, and Digital Persona. 

When using the LivDet 2017 dataset to evaluate algorithm performance, two common experimental protocols, cross-sensor and cross-material, are used to evaluate the generalization performance of the algorithm. Sensors are hardware devices used to collect fingerprint images. As shown in Table \ref{tab.livdet2017}, there are three sensors in the LivDet 2017 dataset: GreenBit, DigitalPersona and Orcanthus. For each sensor, the materials for spoof fingerprints in the training set and testing set are completely different. In the training set, the spoof materials are Wood Glue, Exoflex and Body Double, while the spoof materials in the testing set are Gelatin, Liquid Econflex and Latex. Therefore, in cross-material setting, the sensors used for the training and testing sets are identical, and the only difference between the training set and testing set is the materials used for the spoof fingerprints. In the case of cross-sensor, the training and testing sets not only use different sensors, but also use different spoof materials. The spoof materials used for training set and testing set in the case of cross-sensor are exactly the same as the case of cross-material. The experimental results of the cross-material are shown in Table \ref{tab.crossmaterial}. In the case of the cross-sensor, the sensors used for training and testing are different. Because we have a total of 3 sensors, a total of 6 sets of experiments are performed, and the experimental results are shown in Table \ref{tab.crosssensor}.

\begin{table*}[]
\centering
\caption{SUMMARY OF THE LIVDET 2017 DATASETS \cite{mura2018livdet} }
\label{tab.livdet2017}
\setlength\tabcolsep{10pt}
\huge
\resizebox{.98\textwidth}{!}{%
\begin{tabular}{c|cc|cc|c|c|c|c|c|c}
\hline
\multirow{2}{*}{Sensor} & \multicolumn{4}{c|}{Images}                                                    & \multirow{2}{*}{Model} & \multirow{2}{*}{Type} & \multirow{2}{*}{Resolution(dpi)} & \multirow{2}{*}{Image Size} & \multirow{2}{*}{\begin{tabular}[c]{@{}c@{}}Spoof Material\\ for Training\end{tabular}}       & \multirow{2}{*}{\begin{tabular}[c]{@{}c@{}}Spoof Material\\ for Testing\end{tabular}}        \\ \cline{2-5}
                        & \multicolumn{2}{c|}{Train(Live/Spoof)} & \multicolumn{2}{c|}{Test(Live/Spoof)} &                        &                       &                                  &                             &                                                                                              &                                                                                              \\ \hline
GreenBit                & 1000               & 1200              & 1700              & 2040              & DactyScan84c           & Optical               & 500                              & 500 × 500                     & \multirow{3}{*}{\begin{tabular}[c]{@{}c@{}}Wood Glue,\\ Exoflex,\\ Body Double\end{tabular}} & \multirow{3}{*}{\begin{tabular}[c]{@{}c@{}}Gelatin,\\ Liquid Ecoflex,\\ Latex\end{tabular}} \\ \cline{1-9}
DigitalPersona          & 999                & 1199              & 1700              & 2028              & U.are.U 5160           & Optical               & 500                              & 252 × 324                     &                                                                                              &                                                                                              \\ \cline{1-9}
Orcanthus               & 1000               & 1200              & 1700              & 2018              & Certis2 Imag           & Thermal swipe         & 500                              & 300 × n                       &                                                                                              &                                                                                              \\ \hline
\end{tabular}%
}
\end{table*}

To promote network learning of nonlocal features, we use the cut-out module to perform the data preprocessing. Inspired by \cite{devries2017improved}, we randomly select a pixel coordinate within the image as a center point and then place the zero-mask around that location in the training stage. The size of the zero-mask is set to 96×96, and the number of zero-masks is 10. In addition, data augmentation techniques, including horizontal flipping, vertical flipping, and random rotation, are used to ensure that the trained model is robust to the possible variations in fingerprint images.

MobileNet v2 \cite{2018MobileNetV2} pretrained on ImageNet \cite{russakovsky2015imagenet} is used as the backbone for the proposed CFD-PAD method. The feature generator $G(\cdot)$, feature embedding $E(\cdot)$, and classifier $C(\cdot)$, are all substructures of MobileNet v2. As shown in Table \ref{fig.mobilenetv2}, feature embedding $E(\cdot)$ contains a bottleneck layer and a conv2d 1$\times$1 layer. Compared with other popular deep networks, MobileNet v2 is particularly suitable for mobile applications, allows memory-efficient inference, and only involves standard operations, which was useful for fingerprint PAD in \cite{8987320}. Based on network architecture search (NAS), MobileNet v2 uses a combination of depth-wise separable convolutions \cite{zhang2018shufflenet,howard2017mobilenets,chollet2017xception}, linear bottlenecks, inverted residuals, and information flow interpretation as building blocks. The architecture of linear bottlenecks captures the low-dimensional manifold of interest by inserting linear bottleneck layers into the convolutional blocks and uses linear layers to prevent nonlinearities from destroying too much information. In the inverted residual, which is considered a bottleneck, Sandler \cite{2018MobileNetV2} et al. use shortcuts directly between the bottlenecks because the bottlenecks contain all the necessary information, while an expansion layer merely acts as an implementation detail that accompanies a nonlinear transformation of the tensor. As listed in TABLE \ref{fig.mobilenetv2}, the backbone used in this paper is called MobileNet v2, which consists of 17 bottlenecks, 2 convolutional layers and 1 average pooling layer. Because the sizes of fingerprint images derived from the various sensors are different, the average pooling layer (i.e., avgpool in TABLE \ref{fig.mobilenetv2}) is used to manage arbitrary sizes of input fingerprint images. In addition, the last layer of the original architecture, which is a convolutional layer with 1000-unit output, is replaced by the convolutional layer with a 2-unit output of fingerprint PAD (i.e., live fingerprint and spoof fingerprint).

The implementation used in this study is based on the public platform PyTorch \cite{paszke2017automatic}. We use the Adam optimizer for training, and the learning rate of Adam is set to 0.0001. The feature map $f$ derived after the feature generator is shown in Table \ref{fig.mobilenetv2} and marked as $^*$, and the shape of $f$ is $(7 \times 7 \times 160)$ if the input $x$ is $(224 \times 224 \times 3)$. The feature map $f$ has a total of 160 channels, and the number of important channels $k$ is set to 30. The proposed experiments are conducted on Tesla v100 and TITAN XP.

\subsection{Evaluation Metrics and Comparison Method}
To evaluate the performance of the methods, six popular metrics like average classification error (ACE), true detection rate @ false detection rate = 1$\%$ (TDR@FDR=1.0\%), equal error rate (EER), attack presentation classification error rate (APCER), bona fide presentation classification error rate (BPCER) and ACER, are used \cite{standard2017information}. ACE describes the classification performance of the method. The smaller the value of ACE, the better the performance of the evaluation method. TDR@FDR=1$\%$ represents the percentage of PAs able to breach biometric system security when the reject rate of legitimate users is $\le$ 1\%. APCER describes the proportion of attack presentations using the same presentation attack instrument species incorrectly classified as bona fide presentations in a specific scenario. BPCER is the proportion of bona fide presentations incorrectly classified as presentation attacks in a specific scenario. The ACER can be formulated as:
\begin{equation}
    ACER = \frac{APCER + BPCER}{2}
    \label{eq:acer}
\end{equation}
To demonstrate the performance of the proposed method, we compare it with single-model and multiple-model-based methods. The single-model-based methods include the winner of LivDet 2017 \cite{mura2018livdet} and FSB \cite{chugh2018fingerprint}. Similarly, we compared the proposed method with multiple-model-based methods, i.e., FSB + UMG Wrapper \cite{chugh2020fingerprint} and RTK-PAD \cite{liu2021fingerprint}.

\begin{table*}[]
\caption{Performances of CFD-PAD with or without Each Proposed Module In Terms of ACE and TDR@FDR=1.0\%}
\label{tab.ablation}
\setlength\tabcolsep{15pt}
\huge
\resizebox{.98\textwidth}{!}{
\begin{tabular}{cccc|cc|cc|cc|cc}
\hline
                           &                                                                               &                                                                                                &                                                                                                     & \multicolumn{8}{c}{LivDet 2017}                                                                                                                           \\ \cline{5-12} 
                           &                                                                               &                                                                                                &                                                                                                     & \multicolumn{2}{c|}{GreenBit}     & \multicolumn{2}{c|}{Orcanthus}    & \multicolumn{2}{c|}{DigitalPersona} & \multicolumn{2}{c}{Mean}                     \\ \cline{5-12}
\multirow{-3}{*}{Baseline} & \multirow{-3}{*}{\begin{tabular}[c]{@{}c@{}}PA-Adaptation\\  Loss\end{tabular}} & \multirow{-3}{*}{\begin{tabular}[c]{@{}c@{}}Channel-wise \\ Feature \\ Denoising\end{tabular}} & \multirow{-3}{*}{\begin{tabular}[c]{@{}c@{}}Channel-wise \\ Feature \\ Regularization\end{tabular}} & ACE(\%)       & TDR(\%) & ACE(\%)       & TDR(\%) & ACE(\%)        & TDR(\%)  & ACE(\%)              & TDR(\%)     \\ \hline
$\checkmark$               & $\times$                                                                      & $\times$                                                                                       & $\times$                                                                                            & 3.54          & 81.67             & 2.94          & 71.46             & 4.83           & 68.64              & 3.77 ± 0.97          & 73.92 ± 6.86          \\
$\checkmark$               & $\checkmark$                                                                  & $\times$                                                                                       & $\times$                                                                                            & 2.93          & 88.08             & 2.67          & 82.61             & 4.72           & 76.53              & 3.44 ± 1.12          & 82.41 ± 5.78          \\
$\checkmark$               & $\times$                                                                      & $\checkmark$                                                                                   & $\times$                                                                                            & 3.36          & 86.72             & 2.27          & 84.59             & 4.61           & 71.80              & 3.41 ± 1.17          & 81.04 ± 8.07          \\
$\checkmark$               & $\checkmark$                                                                  & $\times$                                                                                       & $\checkmark$                                                                                        & 3.00          & 90.81             & 2.60          & 86.67             & 4.79           & 76.33              & 3.46 ± 1.16          & 84.60 ± 7.46          \\ \hline
\rowcolor[HTML]{C0C0C0} 
$\checkmark$               & $\checkmark$                                                                  & $\checkmark$                                                                                   & $\times$                                                                                            & \textbf{2.59} & \textbf{93.43}    & \textbf{1.68} & \textbf{97.32}    & \textbf{3.33}  & \textbf{90.73}      & \textbf{2.53 ± 0.82} & \textbf{93.83 ± 3.31} \\ \hline
\end{tabular}
}
\end{table*}

\begin{table*}[]
\centering
\caption{Performance Comparison between the proposed method and state-of-the-art results reported on LivDet2017 dataset for cross material experiments In Terms of ACE and TDR@FDR=1.0\%}
\label{tab.crossmaterial}
\setlength\tabcolsep{11pt}
\huge
\resizebox{.98\textwidth}{!}{%
\begin{tabular}{c|clcccc|cccc}
\hline
                              & \multicolumn{6}{c|}{Single Model}                                                                                                                                                                     & \multicolumn{4}{c}{Multiple Model}                                                     \\ \cline{2-11} 
                              & \multicolumn{2}{c|}{LivDet 2017 Winner \cite{mura2018livdet} } & \multicolumn{2}{c|}{FSB  \cite{chugh2018fingerprint}}                                                  & \multicolumn{2}{c|}{\cellcolor[HTML]{C0C0C0}CFD-PAD (Ours)}                     & \multicolumn{2}{c|}{FSB + UMG \cite{chugh2020fingerprint}}                       & \multicolumn{2}{c}{RTK-PAD \cite{liu2021fingerprint}}     \\ \cline{2-11} 
\multirow{-3}{*}{LivDet 2017} & \multicolumn{2}{c|}{ACE(\%)}            & \multicolumn{1}{c|}{ACE(\%)}     & \multicolumn{1}{c|}{TDR(\%)} & \cellcolor[HTML]{C0C0C0}ACE(\%)     & \cellcolor[HTML]{C0C0C0}TDR(\%) & ACE(\%)     & \multicolumn{1}{c|}{TDR(\%)} & ACE(\%)     & TDR(\%) \\ \hline
GreenBit                      & \multicolumn{2}{c|}{3.56}               & \multicolumn{1}{c|}{3.32}        & \multicolumn{1}{c|}{91.07}             & \cellcolor[HTML]{C0C0C0}\textbf{2.59}        & \cellcolor[HTML]{C0C0C0}\textbf{93.43}             & 2.58        & \multicolumn{1}{c|}{92.29}             & 1.92        & 96.82             \\
Digital Persona               & \multicolumn{2}{c|}{6.29}               & \multicolumn{1}{c|}{4.88}        & \multicolumn{1}{c|}{62.29}             & \cellcolor[HTML]{C0C0C0}\textbf{3.33}        & \cellcolor[HTML]{C0C0C0}\textbf{90.73}             & 4.80        & \multicolumn{1}{c|}{75.47}             & 3.25        & 80.57             \\
Orcanthus                     & \multicolumn{2}{c|}{4.41}               & \multicolumn{1}{c|}{5.49}        & \multicolumn{1}{c|}{66.59}             & \cellcolor[HTML]{C0C0C0}\textbf{1.68}        & \cellcolor[HTML]{C0C0C0}\textbf{97.32}             & 4.99        & \multicolumn{1}{c|}{74.45}             & 1.67        & 96.18             \\ \hline
Mean ± s.d.                   & \multicolumn{2}{c|}{4.75 ± 1.40}        & \multicolumn{1}{c|}{4.56 ± 1.12} & \multicolumn{1}{c|}{73.32 ± 15.52}     & \cellcolor[HTML]{C0C0C0}\textbf{2.53 ± 0.82} & \cellcolor[HTML]{C0C0C0}\textbf{93.83 ± 3.31}      & 4.12 ± 1.34 & \multicolumn{1}{c|}{80.74 ± 10.02}     & 2.28 ± 0.69 & 91.19 ± 7.51      \\ \hline
\end{tabular}%
}
\end{table*}

\begin{table*}[]
\centering
\caption{Performance Comparison between the proposed method and state-of-the-art results reported on LivDet2017 dataset for cross-sensor experiments In Terms of ACE and TDR@FDR=1.0\%}
\label{tab.crosssensor}
\setlength\tabcolsep{30pt}
\huge
\resizebox{\textwidth}{!}{%
\begin{tabular}{cc|cccccc}
\hline
\multicolumn{2}{c|}{}                              & \multicolumn{4}{c|}{Single Model}                                                                                                                               & \multicolumn{2}{c}{Multiple Model} \\ \cline{3-8} 
\multicolumn{2}{c|}{\multirow{-2}{*}{LivDet 2017}} & \multicolumn{2}{c|}{FSB \cite{chugh2018fingerprint}}                               & \multicolumn{2}{c|}{\cellcolor[HTML]{C0C0C0}CFD-PAD(Ours)}                                             & \multicolumn{2}{c}{RPK-PAD \cite{liu2021fingerprint}}        \\ \hline
Training Sensor          & Testing Sensor          & ACE(\%)       & \multicolumn{1}{c|}{TDR(\%)} & \cellcolor[HTML]{C0C0C0}ACE(\%)       & \multicolumn{1}{c|}{\cellcolor[HTML]{C0C0C0}TDR(\%)} & ACE(\%)        & TDR(\%) \\ \hline
GreenBit                 & Orcanthus               & 50.57         & \multicolumn{1}{c|}{0.00}              & \cellcolor[HTML]{C0C0C0}\textbf{22.34}         & \multicolumn{1}{c|}{\cellcolor[HTML]{C0C0C0}\textbf{21.56}}             & 30.49          & 20.61             \\
GreenBit                 & DigitalPersona          & 10.63         & \multicolumn{1}{c|}{57.48}             & \cellcolor[HTML]{C0C0C0}\textbf{7.03}          & \multicolumn{1}{c|}{\cellcolor[HTML]{C0C0C0}\textbf{75.05}}             & 7.41           & 70.41             \\
Orcanthus                & GreenBit                & \textbf{30.07}         & \multicolumn{1}{c|}{\textbf{8.02}}              & \cellcolor[HTML]{C0C0C0}30.92         & \multicolumn{1}{c|}{\cellcolor[HTML]{C0C0C0}2.87}              & 28.81          & 15.00             \\
Orcanthus                & DigitalPersona          & 42.01         & \multicolumn{1}{c|}{\textbf{4.97}}              & \cellcolor[HTML]{C0C0C0}\textbf{27.44}         & \multicolumn{1}{c|}{\cellcolor[HTML]{C0C0C0}4.78}              & 29.19          & 13.26             \\
DigitalPersona           & GreenBit                & 10.46         & \multicolumn{1}{c|}{57.06}             & \cellcolor[HTML]{C0C0C0}\textbf{6.23}          & \multicolumn{1}{c|}{\cellcolor[HTML]{C0C0C0}\textbf{66.57}}             & 6.74           & 70.25             \\
DigitalPersona           & Orcanthus               & 50.68         & \multicolumn{1}{c|}{0.00}              & \cellcolor[HTML]{C0C0C0}\textbf{24.83}         & \multicolumn{1}{c|}{\cellcolor[HTML]{C0C0C0}\textbf{23.09}}             & 28.55          & 18.68             \\ \hline
\multicolumn{2}{c|}{Mean ± s.d.}                   & 32.40 ± 18.53 & \multicolumn{1}{c|}{21.26 ± 28.06}     & \cellcolor[HTML]{C0C0C0}\textbf{19.80 ± 10.59} & \cellcolor[HTML]{C0C0C0}\textbf{32.32 ± 31.07}                          & 21.87 ± 10.48  & 34.70 ± 25.30     \\ \hline
\end{tabular}%
}
\end{table*}

\begin{table*}[]
\centering
\caption{More details of the results in the case of cross-material and cross-sensor.}
\label{tab.result}
\setlength\tabcolsep{11pt}
\huge
\resizebox{.98\textwidth}{!}{%
\begin{tabular}{c|cc|cccccc}
\hline
LivDet 2017                     & Training Sensor & Testing Sensor & ACE(\%)       & TDR@FDR=1.0\%(\%) & EER           & APCER        & BPCER         & ACER         \\ \hline
\multirow{4}{*}{cross-material} & GreenBit        & GreenBit       & 2.59          & 93.43             & 2.82          & 2.12         & 3.05          & 2.59         \\
                                & Orcanthus       & Orcanthus      & 1.68          & 97.32             & 1.83          & 1.59         & 1.76          & 1.68         \\
                                & DigitalPersona  & DigitalPersona & 3.33          & 90.73             & 3.36          & 3.40         & 3.25          & 3.33         \\ \cline{2-9} 
                                & \multicolumn{2}{c|}{Mean ± s.d.} & 2.53 ± 0.83   & 93.83 ± 3.31      & 2.67 ± 0.78   & 2.37 ± 0.93  & 2.69 ± 0.81   & 2.53 ± 0.83  \\ \hline
\multirow{7}{*}{cross-sensor}   & GreenBit        & Orcanthus      & 22.34         & 21.56             & 22.48         & 23.09        & 21.59         & 22.34        \\
                                & GreenBit        & DigitalPersona & 7.03          & 75.05             & 7.26          & 9.52         & 4.55          & 7.04         \\
                                & Orcanthus       & GreenBit       & 30.92         & 2.87              & 33.45         & 16.77        & 45.08         & 30.93        \\
                                & Orcanthus       & DigitalPersona & 27.44         & 4.78              & 28.58         & 20.07        & 34.81         & 27.44        \\
                                & DigitalPersona  & GreenBit       & 6.23          & 66.57             & 6.51          & 4.60         & 7.85          & 6.23         \\
                                & DigitalPersona  & Orcanthus      & 24.83         & 23.09             & 25.78         & 22.60        & 27.06         & 24.83        \\ \cline{2-9} 
                                & \multicolumn{2}{c|}{Mean ± s.d.} & 19.80 ± 10.59 & 32.32 ± 31.07     & 20.68 ± 11.27 & 16.11 ± 7.52 & 23.49 ± 15.58 & 19.8 ± 10.59 \\ \hline
\end{tabular}%
}
\end{table*}

\begin{figure}[!]
    \centering
    \includegraphics[scale=0.6]{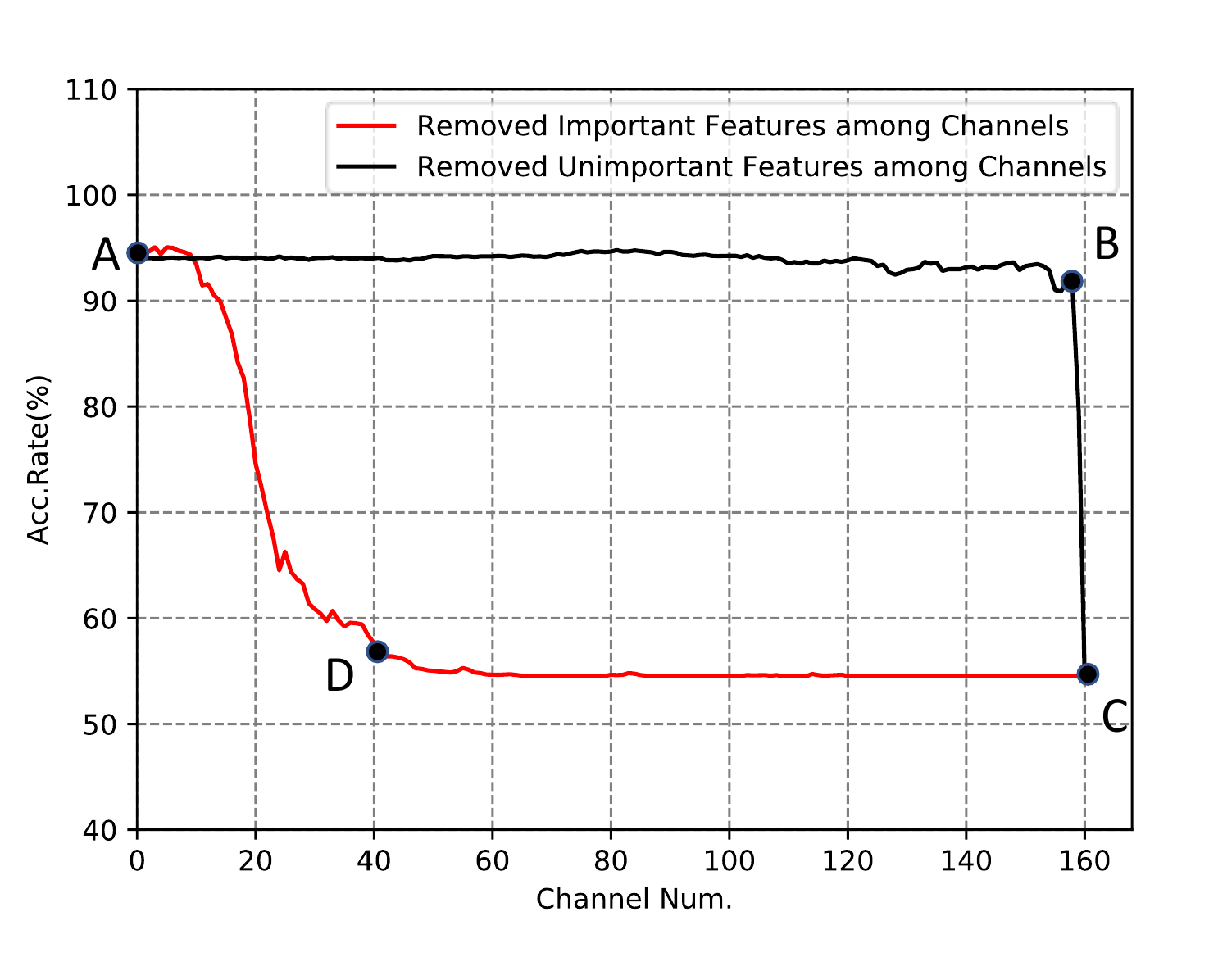}
   \caption{Accuracy after removing different channels. The black line in the figure shows the accuracy after removing "noise" features among channels. The red line in the figure shows the accuracy after removing important features among channels.}
   \label{fig:FigAcc}
\end{figure}

\subsection{Effectiveness Validation of the Proposed CFD-PAD Method}
To verify the effectiveness of the proposed method, we trained a CNN for PAD and derived a feature map at the middle layer of the trained CNN. The derived feature map has a total of 160 channels, and many features extracted from discriminative regions and noise regions are preserved in the feature map. As shown in Fig. \ref{fig:FigAcc}, when some important features or "noise" features among channels are removed, the accuracy of PAD become different. The red line in the figure shows the accuracy after removing important features among channels, and the black line shows the accuracy after removing "noise" features among channels. When approximately 20 important features among channels are removed, PAD accuracy drops from over 90 to less than 80, which is a remarkable decline. After removing nearly 40 important channels, the network can hardly make any correct prediction. In contrast, the changes in accuracy for removing "noise" features among channels are different. As shown by the black line, the accuracy after removal of the first 150 "noise" features among channels does not change significantly. The classification accuracy begins to drop until more than 150 "noise" features among channels are removed. These experimental results show that "noise" and important features exist in the feature map simultaneously. The channels in the feature map that contain abundant discriminative features can be deemed important channels, while the remaining channels are "noise" channels. The discriminative features correspond to those important channels between points A and B on the red line in Fig. \ref{fig:FigAcc}. In contrast, the "noise" channels in Fig. \ref{fig:FigAcc}, which correspond to point A to D on the black line, are similar to the "noise" features. "Noise" channels contain many invalid and redundant "noise" features, which will strongly interfere with the prediction results of the network. Successful removal of the influence of "noise" features improves the performance of the model on PAD.

\begin{figure*}[!]
    \centering
        \includegraphics[scale=0.40]{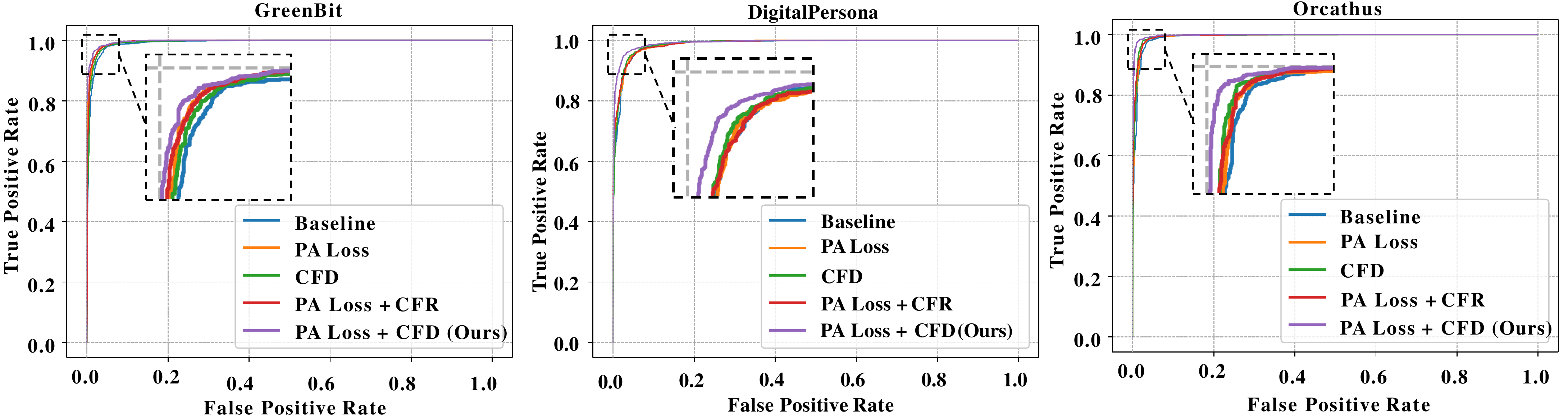}
  \caption{ROC curves for the ablation study on fingerprint PAD. PA Loss denotes the PA-Adaptation loss, CFD denotes the channel-wise feature denoising and CFR denotes the channel-wise feature regularization.}
    \label{fig:roc}
\end{figure*}


We also perform an ablation study to evaluate the cross-material performance gained by each module for different network architectures (i.e., PA-Adaptation loss and channel-wise feature denoising on LivDet 2017). Table \ref{tab.ablation} and Fig. \ref{fig:roc} present the generalizabiliby of the proposed method when all the training and testing images are captured using the same sensor, but the spoof materials in the training and testing sets are completely different. Table \ref{tab.ablation} shows the ablation study results of the proposed method. The PA-Adaptation loss module and channel-wise feature denoising module can effectively improve performance. To visualize the ablation study, we plot the ROC curves for GreenBit, DigitalPersona, and Orcathus in Fig.\ref{fig:roc}. As shown in the figure, the "PA Loss + CFD(Ours)" line achieved the best results in all three cases, which indicates the effectiveness of the proposed method.

To quantify the contribution of PA-Adaptation loss, we compare the baseline with and without PA-Adaptation loss. The Baseline in Table \ref{tab.ablation} is a simple classification network (MobileNet v2), which uses the weights pretrained on ImageNet, for classification, without using PA-Adaption Loss, Channel-wise Feature Denoising and Channel-wise Feature Regularization. Table \ref{tab.ablation} shows the result after adding the PA-Adaptation loss to baseline: ACE decreased from 3.77\% to 3.44\%, and TDR@FDR=1.0\% improved from 73.92\% to 82.41\%. In contrast, after removing PA-Adaptation loss on "Channel-wise Feature Denoising + PA-Adaptation Loss", ACE increased from 2.53\% to 3.41\%, and TDR@FDR=1.0\% declined from 91.68\% to 81.04\%. The results indicate that PA-Adaptation loss has successfully modified the feature distribution of live and spoof fingerprints by clustering all the live fingerprints and pushing different spoof attacks apart. This result highlights that PA-Adaptation loss can effectively improve the performance of the baseline.

Because the channel-wise feature denoising module is an important part of the method, we also test the contribution of the channel-wise feature denoising module, and the results are shown in Table \ref{tab.ablation}. Compared with the baseline, the method using channel-wise feature denoising achieves better performance; ACE decreases from 3.77\% to 3.41\% and TDR@FDR=1.0\% increases from 73.92\% to 81.04\%.
However, when the channel-wise feature denoising module is removed from the final experiment, the ACE increases from 2.53\% to 3.44\%, and TDR@FDR=1.0\% decreases from 93.83\% to 82.41\%. These results demonstrate that the channel-wise feature denoising module can remove redundant "noise" features in the fingerprint effectively.

We also compare the proposed method with channel-wise feature regularization. Channel-wise feature regularization and channel-wise denoising are different in how they suppress channels, but the process of evaluating each channel is the same. If the network only leverages the most discriminative regions while neglecting other auxiliary regions when making a prediction, it may lead to insufficient feature diversity and result in a feature space with low dimensions. To improve network robustness, a regularization-based method is used. Different from channel-wise feature denoising, channel-wise feature regularization makes the network pay more attention to those auxiliary features while neglecting those discriminative features \cite{liu2021group}. Thus, channel-wise feature regularization adopts the same method as channel-wise feature denoising to evaluate each channel's importance. However, when suppressing channels, channel-wise feature regularization keeps those unimportant $k$ channels unchanged, and suppresses the propagation of other channels. The result in Table \ref{tab.ablation} shows that channel-wise feature regularization achieves an ACE of 3.46\% and TDR@FDR=1.0\% of 84.60\%, which did not achieve as good performance as channel-wise feature denoising. This result demonstrates that compared with channel-wise feature regularization, channel-wise feature denoising is more useful and can effectively improve performance.

\begin{figure}[!]
    \centering
        \includegraphics[scale=0.60]{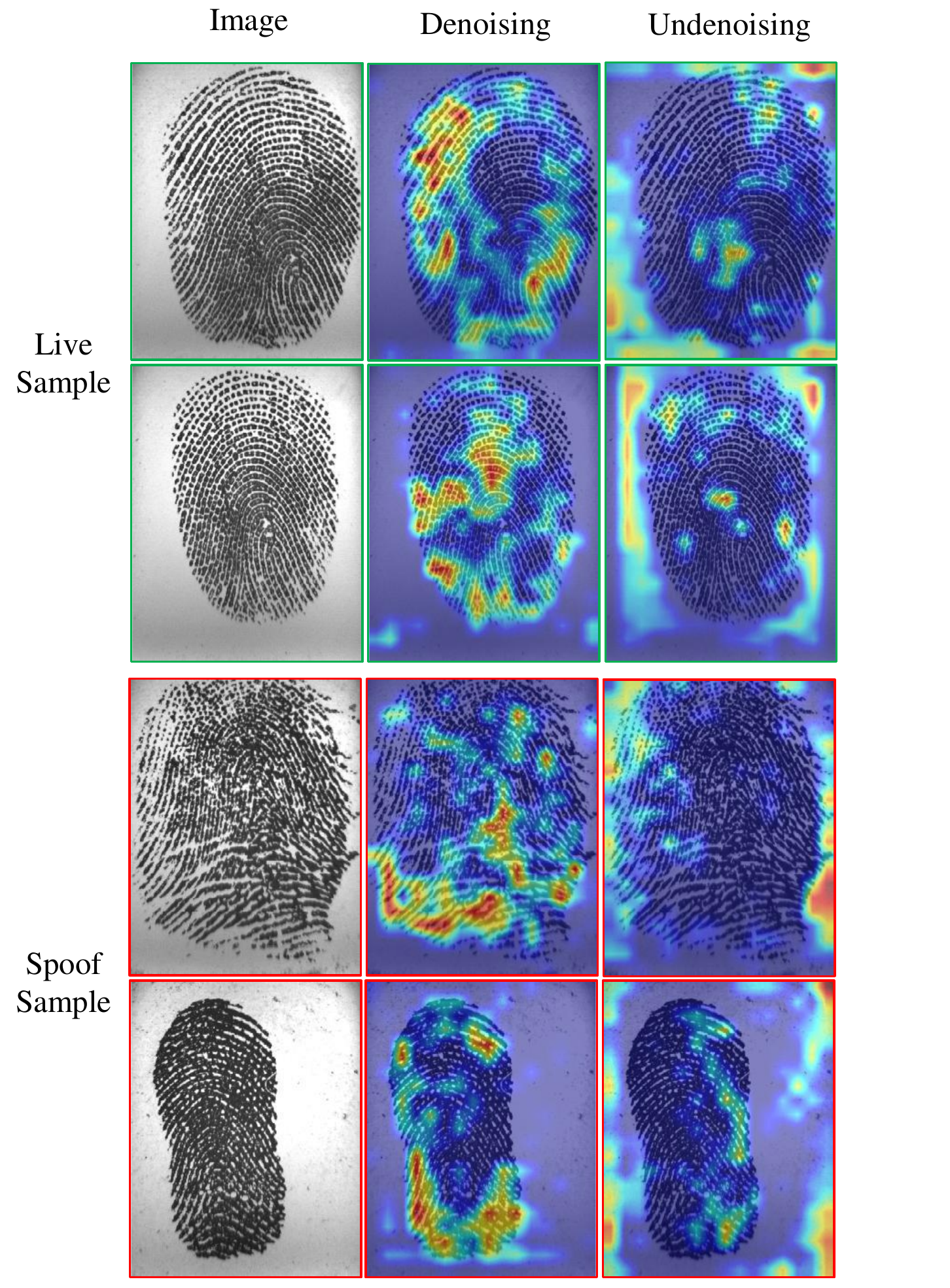}
   \caption{Grad-CAM visualizations on LivDet2017 under Digital Persona. The first two row shows the live sample and the last two row presents the spoof sample. The first column is the original image, the second column is the visualization results after using the denoising method, and the last column is the visualization result without using the denoising method.}
    \label{fig:viscam}
\end{figure}

To investigate the advantage of the proposed method, we use Grad-CAM to provide class activation map (CAM) visualizations of the proposed method and baseline, which has the same architecture as the proposed method and does not involve a denoising module. As shown in Fig. \ref{fig:viscam}, the first two rows show the live sample, and the last two rows present the spoof sample. The first column is the original image, the second column is the visualization results after using the denoising method, and the last column is the visualization result without using the denoising method (undenoising method). CFD-PAD always focuses on the adequate region of interest (ROI) for both live and spoof fingerprints to seek discriminative cues instead of domain-specific backgrounds, which is more likely to generalize well to unseen domains. 
Almost all activated regions are clustered on the fingerprint using denoising method. However, it can be seen that when compared with undenosing method, some activated regions are in background, which indicates the effectiveness of the proposed denoising method.

\subsection{Comparison With Existing Methods}
To demonstrate the performance of the proposed CFD-PAD method, we compare it with other single and multiple-model-based PAD methods, including the winner of LivDet 2017 \cite{mura2018livdet}, a minutiae-centered patch-based method \cite{chugh2018fingerprint}, a style-transfer-based method \cite{chugh2020fingerprint}, and the state-of-the-art PAD method \cite{liu2021fingerprint}. Tables \ref{tab.crossmaterial} and \ref{tab.crosssensor} describe the performances of some fingerprint PAD methods on LivDet 2017. The performance of the proposed CFD-PAD method achieves 2.53\% ACE and 93.83\% TDR@FDR=1\% in the cross-material setting and achieves 19.80\% ACE and 32.32\% TDR@FDR=1\% in the cross-material setting.

In this paper, two different experimental settings, including cross-material and cross-sensor, are considered. In the case of cross-material, the spoof materials used in the training set and testing set are different, as shown in Table \ref{tab.livdet2017}. The spoof materials in the training set are Wood Glue, Ecoflex, Body Double but Gelatin, Latex, and Liquid Ecoflex in the testing set. Therefore, the data partition of LivDet 2017 can be directly used as the cross-material setting. In the case of the cross-sensor, the fingerprint images in the training set and testing set are captured use different sensors.

\subsection{Cross-Material}
\ 
\\
\indent Table \ref{tab.crossmaterial} shows the generalization capability of the proposed method on the LivDet2017 dataset in the case of cross-material. A significant reduction in the ACE is achieved by the proposed method. 
Compared with the best single-model-based method Fingerprint Spoof Buster (FSB) \cite{chugh2018fingerprint}, our approach decreases the mean ACE from 4.56\% to 2.53\%, and increase the mean TDR@FDR=1\% from 73.32\% to 93.83\%. A 44.52\% reduction in the mean ACE and a 27.97\% rise in the mean TDR@FDR=1\% were achieved by the proposed method compared with FSB. Compared with the best multiple-model-based model RTK-PAD \cite{liu2021fingerprint}, the mean of TDR@FDR=1\% increases from 91.19\% to 93.83\%, and the proposed method achieves higher TDR@FDR=1\%.

In the case of Orcanthus, the proposed method outperforms the combination of fingerprint spoof buster and UMG wrapper by a wide margin. When using Orcanthus as the training and testing sensor, the proposed method achieves 1.68\% ACE, which is a significant improvement over the previous methods \cite{mura2018livdet,chugh2018fingerprint,chugh2020fingerprint}. 
When Orcanthus is used as the sensor, the proposed method achieves an ACE 2.73\% lower than the LivDet 2017 winner \cite{mura2018livdet}, which is the best single-model-based method in the case of Orcanthus. Given Digital Persona as the sensor, the ACE of the proposed method is 3.33\%, which is also better than the best single-model-based method \cite{chugh2020fingerprint}. 
Compared with the single-model-based methods, the proposed method outperforms all the single-model-based methods by a wide margin in terms of both ACE and TDR@FDR=1\%.
Compared with multiple-model-based-methods, our performance is still comparable, or better. Even if we use a single-model, we can fully surpass the combination of fingerprint spoof buster and the style-transfer-based wrapper (FSB + UMG) \cite{chugh2020fingerprint} in both ACE and TDR@FDR=1\%.
In terms of TDR@FDR=1\%, the proposed method improved the performance of RTK-PAD \cite{liu2021fingerprint} from 91.19\% to 93.83\%. Experimental results indicate that the proposed CFD-PAD method achieves better performance and generalization capability in PAD, and has achieved the state-of-the-art performance on the LivDet 2017 dataset, in terms of TDR@FDR=1\%.

\begin{table}[]
\centering
\caption{Experimental results with different number of important channels.}
\label{tab.masknum}
\setlength\tabcolsep{11pt}
\huge
\resizebox{.49\textwidth}{!}{%
\begin{tabular}{cc|ccc|c}
\hline
\multicolumn{2}{c|}{Num of Important channels}                                        & GreenBit                               & Orcanthus                              & DigitalPersona                         & Mean ± s.d.                                   \\ \hline
\multicolumn{1}{c|}{}                     & ACE(\%)                                   & 2.86                                   & 2.69                                   & 4.63                                   & 3.39 ± 1.07                                   \\
\multicolumn{1}{c|}{}                     & TDR@FDR=1.0\%(\%)                         & 93.03                                  & 87.31                                  & 78.50                                  & 86.28 ± 7.32                                  \\
\multicolumn{1}{c|}{}                     & EER(\%)                                   & 2.93                                   & 2.88                                   & 4.71                                   & 3.51 ± 1.04                                   \\
\multicolumn{1}{c|}{}                     & APCER(\%)                                 & 3.08                                   & 3.02                                   & 4.59                                   & 3.56 ± 0.89                                   \\
\multicolumn{1}{c|}{}                     & BPCER(\%)                                 & 2.64                                   & 2.35                                   & 4.67                                   & 3.22 ± 1.26                                   \\
\multicolumn{1}{c|}{\multirow{-6}{*}{10}} & ACER(\%)                                  & 2.86                                   & 2.69                                   & 4.63                                   & 3.39 ± 1.07                                   \\ \hline
\multicolumn{1}{c|}{}                     & ACE(\%)                                   & 2.61                                   & 2.20                                   & 4.53                                   & 3.11 ± 1.24                                   \\
\multicolumn{1}{c|}{}                     & TDR@FDR=1.0\%(\%)                         & 91.46                                  & 90.54                                  & 81.07                                  & 87.69 ± 5.75                                  \\
\multicolumn{1}{c|}{}                     & EER(\%)                                   & 2.96                                   & 2.23                                   & 4.62                                   & 3.27 ± 1.22                                   \\
\multicolumn{1}{c|}{}                     & APCER(\%)                                 & 1.82                                   & 1.93                                   & 4.68                                   & 2.81 ± 1.62                                   \\
\multicolumn{1}{c|}{}                     & BPCER(\%)                                 & 3.40                                   & 2.47                                   & 4.37                                   & 3.41 ± 0.95                                   \\
\multicolumn{1}{c|}{\multirow{-6}{*}{20}} & ACER(\%)                                  & 2.61                                   & 2.20                                   & 4.53                                   & 3.11 ± 1.24                                   \\ \hline
\multicolumn{1}{c|}{}                     & \cellcolor[HTML]{C0C0C0}ACE(\%)           & \cellcolor[HTML]{C0C0C0}\textbf{2.59}  & \cellcolor[HTML]{C0C0C0}\textbf{1.68}  & \cellcolor[HTML]{C0C0C0}\textbf{3.33}  & \cellcolor[HTML]{C0C0C0}\textbf{2.53 ± 0.83}  \\
\multicolumn{1}{c|}{}                     & \cellcolor[HTML]{C0C0C0}TDR@FDR=1.0\%(\%) & \cellcolor[HTML]{C0C0C0}\textbf{93.43} & \cellcolor[HTML]{C0C0C0}\textbf{97.32} & \cellcolor[HTML]{C0C0C0}\textbf{90.73} & \cellcolor[HTML]{C0C0C0}\textbf{93.83 ± 3.31} \\
\multicolumn{1}{c|}{}                     & \cellcolor[HTML]{C0C0C0}EER(\%)           & \cellcolor[HTML]{C0C0C0}\textbf{2.82}  & \cellcolor[HTML]{C0C0C0}\textbf{1.83}  & \cellcolor[HTML]{C0C0C0}\textbf{3.36}  & \cellcolor[HTML]{C0C0C0}\textbf{2.67 ± 0.78}  \\
\multicolumn{1}{c|}{}                     & \cellcolor[HTML]{C0C0C0}APCER(\%)         & \cellcolor[HTML]{C0C0C0}\textbf{2.12}  & \cellcolor[HTML]{C0C0C0}\textbf{1.59}  & \cellcolor[HTML]{C0C0C0}\textbf{3.40}  & \cellcolor[HTML]{C0C0C0}\textbf{2.37 ± 0.93}  \\
\multicolumn{1}{c|}{}                     & \cellcolor[HTML]{C0C0C0}BPCER(\%)         & \cellcolor[HTML]{C0C0C0}\textbf{3.05}  & \cellcolor[HTML]{C0C0C0}\textbf{1.76}  & \cellcolor[HTML]{C0C0C0}\textbf{3.25}  & \cellcolor[HTML]{C0C0C0}\textbf{2.69 ± 0.81}  \\
\multicolumn{1}{c|}{\multirow{-6}{*}{30}} & \cellcolor[HTML]{C0C0C0}ACER(\%)          & \cellcolor[HTML]{C0C0C0}\textbf{2.59}  & \cellcolor[HTML]{C0C0C0}\textbf{1.68}  & \cellcolor[HTML]{C0C0C0}\textbf{3.33}  & \cellcolor[HTML]{C0C0C0}\textbf{2.53 ± 0.83}  \\ \hline
\multicolumn{1}{c|}{}                     & ACE(\%)                                   & 2.96                                   & 2.50                                   & 3.99                                   & 3.15 ± 0.76                                   \\
\multicolumn{1}{c|}{}                     & TDR@FDR=1.0\%(\%)                         & 92.37                                  & 90.24                                  & 82.50                                  & 88.37 ± 5.19                                  \\
\multicolumn{1}{c|}{}                     & EER(\%)                                   & 3.17                                   & 2.53                                   & 4.30                                   & 3.33 ± 0.9                                    \\
\multicolumn{1}{c|}{}                     & APCER(\%)                                 & 1.77                                   & 2.53                                   & 3.25                                   & 2.52 ± 0.74                                   \\
\multicolumn{1}{c|}{}                     & BPCER(\%)                                 & 4.16                                   & 2.47                                   & 4.73                                   & 3.79 ± 1.18                                   \\
\multicolumn{1}{c|}{\multirow{-6}{*}{40}} & ACER(\%)                                  & 2.97                                   & 2.50                                   & 3.99                                   & 3.15 ± 0.76                                   \\ \hline
\multicolumn{1}{c|}{}                     & ACE(\%)                                   & 3.23                                   & 2.59                                   & 4.12                                   & 3.31 ± 0.77                                   \\
\multicolumn{1}{c|}{}                     & TDR@FDR=1.0\%(\%)                         & 91.06                                  & 86.92                                  & 78.80                                  & 85.59 ± 6.24                                  \\
\multicolumn{1}{c|}{}                     & EER(\%)                                   & 3.45                                   & 2.77                                   & 4.22                                   & 3.48 ± 0.73                                   \\
\multicolumn{1}{c|}{}                     & APCER(\%)                                 & 2.17                                   & 2.13                                   & 4.04                                   & 2.78 ± 1.09                                   \\
\multicolumn{1}{c|}{}                     & BPCER(\%)                                 & 4.28                                   & 3.06                                   & 4.20                                   & 3.85 ± 0.68                                   \\
\multicolumn{1}{c|}{\multirow{-6}{*}{50}} & ACER(\%)                                  & 3.23                                   & 2.60                                   & 4.12                                   & 3.31 ± 0.77                                   \\ \hline
\multicolumn{1}{c|}{}                     & ACE(\%)                                   & 3.27                                   & 2.80                                   & 4.31                                   & 3.46 ± 0.77                                   \\
\multicolumn{1}{c|}{}                     & TDR@FDR=1.0\%(\%)                         & 86.26                                  & 89.00                                  & 75.10                                  & 83.45 ± 7.36                                  \\
\multicolumn{1}{c|}{}                     & EER(\%)                                   & 3.53                                   & 3.12                                   & 4.44                                   & 3.7 ± 0.68                                    \\
\multicolumn{1}{c|}{}                     & APCER(\%)                                 & 2.32                                   & 2.13                                   & 4.24                                   & 2.9 ± 1.17                                    \\
\multicolumn{1}{c|}{}                     & BPCER(\%)                                 & 4.22                                   & 3.47                                   & 4.37                                   & 4.02 ± 0.48                                   \\
\multicolumn{1}{c|}{\multirow{-6}{*}{80}} & ACER(\%)                                  & 3.27                                   & 2.80                                   & 4.31                                   & 3.46 ± 0.77                                   \\ \hline
\end{tabular}%
}
\end{table}

\subsection{Cross-Sensor}
\ 
\\
\indent We now compare the performance of the proposed method with fingerprint spoof buster \cite{chugh2018fingerprint} and RTK-PAD methods \cite{liu2021fingerprint} under the cross-sensor setting. The sensors used in the training set and testing set are completely different. Because there are three types of sensors, we performed 6 sets of experiments to cover all of their combinations. As the results in Table \ref{tab.crosssensor} show, the proposed method achieves the best mean ACE. More specifically, compared with the best single-model-based method \cite{chugh2018fingerprint}, we achieve better performance in both ACE and TDR@FDR.
Compared with the current best multiple-model-based method, i.e., RTK-PAD \cite{liu2021fingerprint}, we still obtain a better performance in terms of ACE (19.80\% vs. 21.87\%).

FSB+UMG \cite{chugh2020fingerprint} is a style-transfer-based method, which uses the UMG framework to augment the CNN-based spoof detector (FSB). UMG significantly improves FSB performance against novel materials while retaining its performance on known materials \cite{chugh2018fingerprint,chugh2020fingerprint}.
However, their method requires the data captured by the target sensor \cite{chugh2020fingerprint}, while we do not require such data for training. It is thus unfair to compare the proposed method with FSB + UMG \cite{chugh2020fingerprint}.
The FSB + UMG wrapper achieved 19.37\% ACE and 43.23\% TDR@FDR=1\% in the cross-sensor setting. 
However, the proposed method reaches an accuracy close to that of FSB + UMG wrapper (e.g., 19.80\% vs. 19.37\% \cite{chugh2020fingerprint} in terms of ACE), without using any image from the target sensor.
Thus, we conclude that even when the data of target sensors is not available, the proposed method can also effectively improve cross-sensor performance.

The comparative results of the proposed experiments are shown in Tables \ref{tab.crossmaterial} and \ref{tab.crosssensor}. Table \ref{tab.crossmaterial} shows the experimental results in cross material setting, and Table \ref{tab.crosssensor} shows the experimental results in cross-sensor setting. The compared methods can be divided into single-model-based methods and multiple-model-based methods. The performance of the proposed method is worse than the state-of-the-art method in some cases. 
However, multiple model fusion approaches like RTK-PAD \cite{liu2021fingerprint}, applies two models to extract local and global features, which are then fused to get the final prediction. The computational complexity of multiple model fusion approaches is thus substantially higher than ours. In contrast, our proposed method is simpler and much more efficient, i.e. a 74.76\% reduction in computation time over state-of-the-art method, is achieved. Computational complexity in terms of time cost are shown in Table \ref{tab:time}.

\subsection{Number of Important Channels}
The number of important channels $k$ is critical because it can determine the features we leverage. To find a precise $k$ to ensure that no important channels are lost, we conducted the following experiments, where $k$ is set to 10, 20, 30, 40, 50 and 80, to evaluate the performance of the proposed method. Experimental results are shown in Table \ref{tab.masknum}. Performance improves when $k$ is increased from 10 to 30. Then, if $k$ is larger than 30, the performance decrease. When $k$ is set to 30, the average ACE, TDR@FDR=1.0\%, EER, APCER, BPCER and ACER are the best. Thus, we can conclude that 30 is the best number of important channels.

\subsection{Computational Complexity Analysis}
To investigate the computational complexity of the proposed method, we calculate the FLOPs (floating point operations) and time cost of different approaches. As shown in Table \ref{tab:time}, both the time cost and FLOPs of the proposed method are less than those of FSB \cite{chugh2018fingerprint} and RTK-PAD \cite{liu2021fingerprint}.
FSB \cite{chugh2018fingerprint} is a single-model-based method that calculate the spoof score of 48 minutiae-centered patches and fush these scores; no such operations are required by the proposed method. Thus, a 208.3 ms time cost can be saved by the proposed method.
RTK-PAD \cite{liu2021fingerprint} is a multiple-model-based method, that uses a global PAD module, a local PAD module, and a rethinking module. As only a single-model is used. The proposed method spends less time than RTK-PAD. When the input size is 1×224×224×3, the proposed method takes 59.9 ms to distinguish spoof from live fingerprints, which is substantially faster than RTK-PAD.
The proposed method is based on a single lightweight model, which require less computation time and is suitable for real-word applications. Compared with RTK-PAD, a 74.76\% reduction in computation time is achieved by the proposed method.
Thus, whether compared with other single-model or multiple-model-based methods, far less computation time is required by the proposed method.

\begin{table}[H]
\centering
\caption{computational complexity in terms of params and time cost.}
\label{tab:time}
\setlength\tabcolsep{12pt}
\huge
\resizebox{.49\textwidth}{!}{%
\begin{tabular}{c|c|c|c}
\hline
               & Input Size   & FLOPs (GMACs) & Time Cost (ms) \\ \hline
FSB \cite{chugh2018fingerprint}            & $\sim$48×224×224×3 & $\sim$27.84         & 268.2          \\
RTK-PAD \cite{liu2021fingerprint}        & 1×224×224×3  & 0.69          & 237.3          \\
\rowcolor[HTML]{C0C0C0} 
CFD-PAD (Ours) & 1×224×224×3  & \textbf{0.31} & \textbf{59.9} \\ \hline
\end{tabular}%
}
\end{table}

\section{CONCLUSION}
Improving the robustness and generalization capability of PAD methods is critical to ensure the security of fingerprint recognition systems. 
This paper thus proposes a channel-wise feature denoising method and a PA-Adaptation loss.
The method contains three key steps during training, including channel's importance evaluation, "noise" channel suppression and channel-wise domain alignment. During channel evaluation, the importance of each channel is quantified according to the change in the results after the channel is suppressed. Then, all the channels in the feature map can be divided into important channels and "noise" channels. The influence of "noise" features will be substantially reduced due to the suppression of the propagation of "noise" channels. Finally, a PA-Adaptation loss is designed to constrain the feature distribution. The effectiveness of the proposed method was demonstrated experimentally using the LivDet 2017 dataset, which is one of the most recent publicly available LivDet datasets. 
Experimental results showed that compared with baseline, PA-Adaptation loss reduced ACE from 3.77\% to 3.44\% and increased TDR@FDR=1\% from 73.92\% to 82.41\%. Similarly, the channel-wise feature denoising module reduced ACE from 3.77\% to 3.41\% and increased TDR@FDR=1\% from 73.92\% to 81.04\%. The combination of channel-wise feature denoising and PA-Adaptation loss can achieve 2.53\% ACE and 93.83\% TDR@FDR=1\%. The effectiveness of the proposed method is also demonstrated by comparision with the state-of-the-art PAD methods under both cross-material and cross-sensor settings.

\section{Acknowledgment}
This work is supported in part by National Natural Science Foundation of China under Grant 62076163 and Grant 91959108; In part by the Shenzhen Fundamental Research Fund under Grant JCYJ20190808163401646; And in part by the Tencent Rhinoceros Birds-Scientific Research Foundation for Young Teachers of Shenzhen University.

{\small
\bibliographystyle{IEEEtran}
\bibliography{egbib}
}

\end{document}